\documentclass[10pt,conference,compsocconf]{IEEEtran}
\usepackage{times}

\usepackage{caption}
\ifCLASSINFOpdf
\else
\fi
\usepackage{lettrine}
\usepackage{graphicx}
\usepackage{subcaption}
\usepackage{lipsum}
\usepackage{algorithm,algpseudocode}
\usepackage{amsmath}
\usepackage{amsfonts}
\usepackage{amssymb}

\usepackage{verbatim}
\usepackage{todonotes}

\begin{document}
\pagenumbering{gobble}
%
\title{\textbf{\Large A Probabilistic Approach for Discovering Daily Human Mobility Patterns with Mobile Data\\[-1.5ex]}}


\author{\IEEEauthorblockN{~\\[-0.4ex]\large Weizhu Qian\\[0.3ex]\normalsize}
\IEEEauthorblockA{\textit{CIAD, Univ. Bourgogne Franche-Comt\'e,} \\
\textit{UTBM, F-90010}\\
Belfort, France \\
Email: weizhu.qian@utbm.fr}
\and
\IEEEauthorblockN{~\\[-0.4ex]\large Fabrice Lauri\\[0.3ex]\normalsize}
\IEEEauthorblockA{\textit{CIAD, Univ. Bourgogne Franche-Comt\'e,} \\
\textit{UTBM, F-90010}\\
Belfort, France \\
Email: fabrice.lauri@utbm.fr}
\and
\IEEEauthorblockN{~\\[-0.4ex]\large Franck Gechter\\[0.3ex]\normalsize}
\IEEEauthorblockA{\textit{CIAD, Univ. Bourgogne Franche-Comt\'e,} \\
\textit{UTBM, F-90010}\\
Belfort, France \\
Email: franck.gechter@utbm.fr}
}


%


\maketitle

\begin{abstract}
Discovering human mobility patterns with geo-location data collected from smartphone users has been a hot research topic in recent years. In this paper, we attempt to discover daily mobile patterns based on GPS data. We view this problem from a probabilistic perspective in order to explore more information from the original GPS data compared to other conventional methods. A non-parameter Bayesian modeling method, Infinite Gaussian Mixture Model, is used to estimate the probability density for the daily mobility. Then, we use Kullback-Leibler divergence as the metrics to measure the similarity of different probability distributions. And combining Infinite Gaussian Mixture Model and  Kullback-Leibler divergence, we derived an automatic clustering algorithm to discover mobility patterns for each individual user without setting the number of clusters in advance. In the experiments, the effectiveness of our method is validated on the real user data collected from different users. The results show that the IGMM-based algorithm outperforms the GMM-based  algorithm. We also test our methods on the dataset with different lengths to discover the minimum data length for discovering mobility patterns.

\end{abstract}



%
\IEEEpeerreviewmaketitle

\section{Introduction} \label{Sec: Introduction}

\lettrine{S}{martphone} devices are equipped with multiple sensors that can record user behavior on the handsets. With the help of a large-scale smartphone usage data, researchers are able to study human behavior in the real world. Since location information is one of the crucial aspects of human behaving, investigating human mobility from mining mobile data has become recently a popular research topic.

Previous research in this filed mainly only focus on discovering the significant places or predicting the transition among the significant places \cite{do2014places}, \cite{baumann2018selecting}, \cite{mcinerney2013modelling}. However, these research neglect the data sampled at the places where one stay for a relatively short time, for instance, in the middle of transitions. As opposed to this point of view, we believe that these data is important for revealing human mobility patterns as well.

To characterize human mobility, we should realize that there are multiple patterns lying in the mobility data even for the same individuals. For example, normally, on workdays, one goes to work or school at daytime while on weekends, he/she may prefer to stay at home. As a result, the mobility on weekends is different from the one on workdays. Therefore, for each user, it is reasonable to depict the trajectory for each day and then discover the common mobility patterns shared by the trajectories of different days.

In our work, the human mobility is recorded by the GPS module embedded on the smartphone devices. It should be emphasized that the GPS data (longitudes and latitudes) is not evenly distributed spatially because one may stay longer at a significant place (i.e, home or workplace/school) than at a less significant place (i.e, restaurants or the roads). Thus, an appropriate description for human mobility is to treat the location of an individual as a set of data points randomly distributed in the space with respect to different probabilities. Moreover, in practice, the data collecting procedure may not be continuous all the time because the GPS module is turned off or does not function sometimes. As a consequence, it arises the issue of data sparsity. In general, the human mobility GPS data has the following properties:
 
\begin{itemize}
    \item The data has latent structures.
    \item The data is not evenly distributed in space.
    \item The data is sparse and noisy. 
\end{itemize}

\begin{figure}[!t]
\centering
\includegraphics[width= \linewidth]{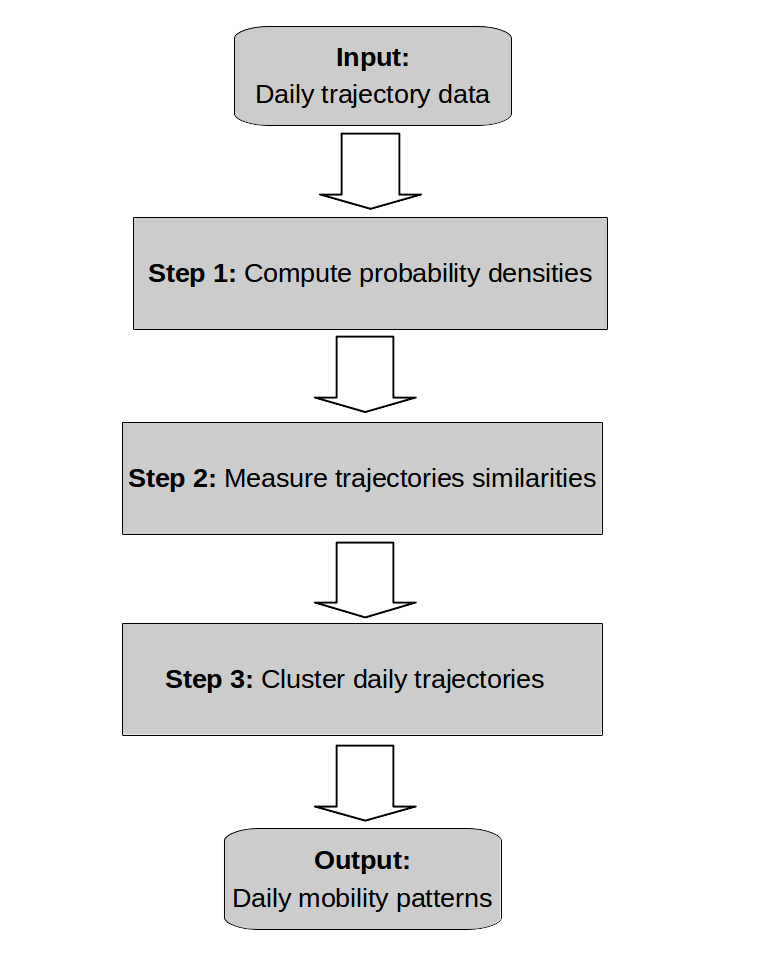}
\caption{Overview of our method.}\label{Fig: Method}
\end{figure}

These unique data characteristics prevent researchers adopting some conventional methods. Therefore, in our work, we adopt a probabilistic approach to describe the daily human mobility. As compared to the conventional methods, we believe our approach can explore more information from the original GPS data and decrease the impact of data sparsity. 
The approach presented in this paper is aimed at investigating this hypothesis and is structured into three steps as shown in Fig.~\ref{Fig: Method}.    


The first stage of the method is to estimate the probability density for each day's trajectories. For such task, Gaussian Mixture Model \cite{reynolds2015gaussian} is a possible solution. However, the standard Gaussian Mixture Model needs to set the number of components in advance, which is not practical to implement because trajectory data can be statistically heterogeneous and a fixed component number for every daily trajectories is not appropriate. To handle this problem, we adopted the Infinite Gaussian Mixture Model \cite{rasmussen2000infinite}, in which the Dirichlet process prior is used to modify the mixed weights of components.

To measure the difference between different mobility probability densities, the Kullback-Leibler divergence \cite{kullback1997information} estimator is used. The Kullback-Leibler divergence is an asymmetric metric, which means the distance from distribution p to distribution q is not the same as the distance from distribution q to distribution p unless they are identical distributions. We exploit the inequality property of KL divergence to reveal the subordinate relationship of one trajectory to another.

Finally, we devise a clustering algorithm using the Infinite Gaussian Mixture Models with Kullback-Leibler divergence to discover the mobility patterns existing in human mobility data. More importantly, as compared to traditional methods, our clustering algorithm is automatic because it does not require a preset of the pattern number.

The work presented in this paper has then 3 main contributions:

\begin{itemize}
    \item For estimating probability density of daily mobility, we illustrate that the Infinite Gaussian Mixture Models outperform the Gaussian Mixture Models.
    \item We prove that Kullback-Leibler divergence is an appropriate metrics to measure the closeness of mobility probability densities.  
    \item We develop a clustering algorithm based on Infinite Gaussian Mixture Model and Kullback-Leibler divergence to find the human mobility patterns.   
\end{itemize}

The reminder of the paper is organized as follows. Section~\ref{Sec:Related Work} surveys the related work. Section~\ref{Sec: Problem Formulation} addresses the problem we are tackling in this paper. In Section~\ref{Sec: Proposed Method}, the proposed method is depicted. In Section~\ref{Sec: Experiments and Results} presents the conducted experiment and its results to evaluate our method with real user data. Finally, we conclude our paper and discuss about the future work in Section~\ref{Sec: Conclusion and Perspective}.

\section{Related Work} \label{Sec:Related Work}

 In literature, previous research such as \cite{lu2013approaching}, \cite{ye2012situation}, \cite{lin2014mining}, \cite{pirozmand2014human}, \cite{zheng2015trajectory}, \cite{cao2007discovery} which were studying human mobility with mobile data, mainly focused on tasks such as extracting significant places, predicting next visiting places, predicting visit duration or clustering trajectories.

A widespread topic is to predict human mobility with the smartphone usage contextual information, e.g., temporal information, application usage, call logs, WiFi status, Cell ID, etc. In \cite{baumann2018selecting} and \cite{do2014and} for instance, the researchers applied various machine learning techniques to accomplish prediction tasks such as next-time slot location prediction and next-place prediction. In particular, they exploited how different combinations of contextual features related to smartphone usage can affect prediction accuracy. Meanwhile, they also compared the predicting performance of individual models and generic models.

Another frequently-used method for such tasks is to use probabilistic models. Through calculating the conditional probabilities between contextual features, \cite{do2012contextual} developed the contextual conditional models for the next-place prediction and visit duration prediction. In \cite{do2015probabilistic} and \cite{peddemors2010predicting}, the researchers presented the probabilistic prediction frameworks based on kernel density estimation. \cite{do2015probabilistic} utilized conditional kernels density estimation to predict the mobility events while \cite{peddemors2010predicting} devised different kernels for different context information types. And in \cite{mcinerney2013modelling}, the authors developed a location Hierarchical Dirichlet Process (HDP) based approach to model heterogeneous location habits under data sparsity.

In addition, generative models can also be applied to predict human mobility. This type of models can be Naïve Bayes \cite{muhlenbrock2004learning}, Markov Model \cite{yu2017modeling}, Hidden Markov Model (HMM) \cite{cho2016exploiting} or Dynamic Bayesian Network (DBN) \cite{etter2013go}, \cite{patterson2003inferring}, etc. These generative models attempt to predict the future states of human behavior by computing the state transition probabilities. However, when the number of states expands, the calculation grows exponentially.

Among the other possible approaches, \cite{zheng2009mining} proposed a Hypertext Induced Topic Search-based inference model for mining interesting locations and travel sequences using a large GPS dataset in certain region. In \cite{do2014places}, the authors employed the random forests classifiers to label different places without any geo-location information. \cite{scellato2011nextplace} made use of nonlinear time series analysis of the arrival time and residence time for location prediction.

In particular, for clustering user trajectories, there exists several different methods. However, these conventional algorithms are not applicable to our objectives. For example, some researchers used K-means \cite{jiang2012clustering}, \cite{ashbrook2003using} in their work, whereas K-means can not handle the trajectories with complex shapes or noisy data because it is based on Euclidean distance. Besides, it also need the pre-knowledge of cluster number, which is not acquirable in many cases.

Though DBSCAN \cite{tang2015uncovering}, \cite{yu2017modeling}, a density based clustering techniques, can deal with data with arbitrary shapes and does not require the number of cluster in advance. However, it still needs to set the minimum points number and neighbourhood radius to recognize core areas and it treats the non-core data points as noise. From our study, we argue that the trajectory parts with less data density are also essential to demonstrate the human mobility trajectories. And the grid searching algorithm \cite{do2012contextual} focus on detecting the stay points within a set of square regions and fails to reveal the mobility at a larger scale.


In this paper we decided to focus on a probabilistic point of view. As compared to aforementioned previous works, we aim at describing the daily trajectories using their probability densities. Moreover, to discover the common mobility patterns shared among these trajectories, we devise an automatic clustering algorithm. As opposed to traditional clustering algorithms, our method is able to exploit more information from the sparse and noisy original GPS data and free from pre-defining clusters number.

\section{Problem Formulation} \label{Sec: Problem Formulation}

As expressed in introduction, our purpose is to discover the mobility patterns for each individual from their GPS location data. 

\begin{figure}[!t]
\centering
\includegraphics[width= \linewidth]{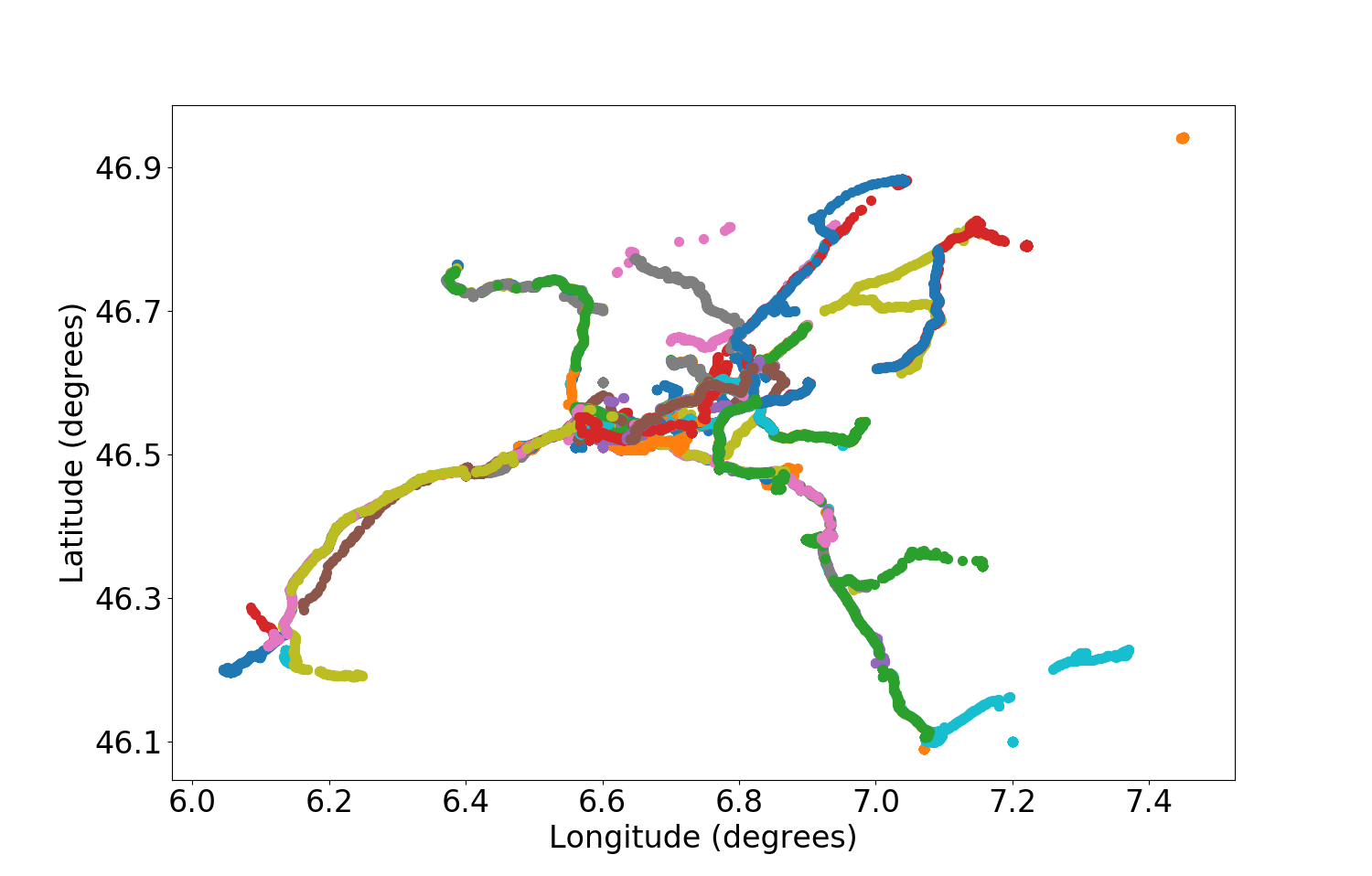}
\caption{GPS data for a randomly selected user. Different colors represent different days.}\label{Fig: GPS_Data}
\end{figure}

As shown in Fig.~\ref{Fig: GPS_Data}, the mobility for one individual consists of many different trajectories (the data is from the MDC dataset, the detailed data description will be in following experiments). We believe that one's daily mobility is rather regular and there are  common mobility patterns shared among different daily trajectories. From a common sense, one may follow the regular daily itineraries, for instance, home-work place/school-home. Yet, on different days the daily itineraries may not be the same, for instance, on the way to home, one may take a detour to do shopping in a supermarket sometimes. Hence, our objective is to discover all the potential daily mobility from the data with location information.      

\begin{figure}[!t]
\centering
\includegraphics[width= \linewidth]{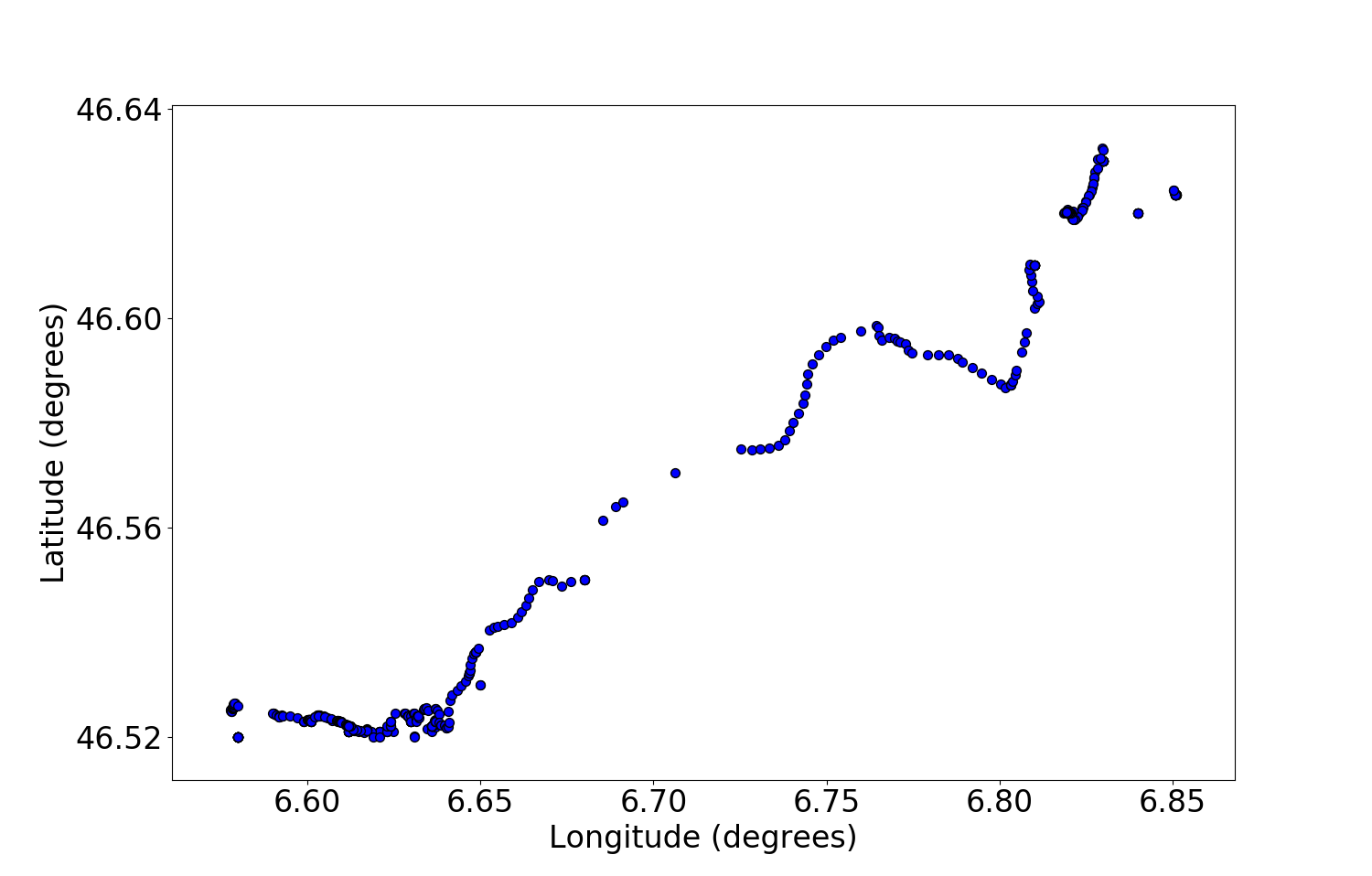}
\caption{One randomly selected daily trajectories for a given user.}\label{Fig: T1}
\end{figure}

We extract each day's trajectory from the whole dataset as shown in Fig.~\ref{Fig: T1}. 
Fig.~\ref{Fig: T1} reveals that a daily trajectories recorded by GPS data is not distributed evenly in space, and is even not continuous in some areas. It may be caused by the data collecting procedure: some data collecting time range is actually relatively short (less than 24 hours, in fact, only few hours sometimes), which leads to the data sparsity problem. 

In order to overcome this problem and exploit as much information as possible from the GPS data, we argue that a reasonable way to describe the daily trajectories is to estimate the probability density of the location data. And the relationship among the trajectories can be represented by their probability densities. As a result, we can discover all the mobility patterns for each user. 

The tasks in this paper will be as follows: 

\begin{itemize}
\item Task 1: Estimate the probability density for mobility for each day. We will compare results of GMM and IGMM. 
\item Task 2: Measure the closeness between different trajectories. We will use the KD divergence as metrics.     
\item Task 3: Discover the similar mobility patterns among all the recorded daily trajectories. This can be regarded as a clustering problem.    
\item Task 4: Compare the IGMM algorithm with the GMM based algorithms.
\item Task 5: Identify the minimum data length for discovering all mobility patterns.
\end{itemize}

\section{Proposed Method} \label{Sec: Proposed Method}

\subsection{Estimate Daily Trajectories Probability Density}

We assume that the GPS location data points are distributed randomly spatially. Besides, the distribution of each day consists of unknown number of heterogeneous sub-distributions. Therefore, it is reasonable to adopt mixed Gaussian models for estimating probability density of daily mobility.  
 
\subsubsection{Gaussian Mixture Model}
A Gaussian Mixture Model (GMM) is composed of a fixed number $K$ of sub components. The probability distribution of a GMM can be described as follows: 

\begin{equation}\label{Equ: GMM}
P(x)=\sum_{k=1}^{K}{\pi_k P(x | \theta_k)}
\end{equation}


where, $x$ is the observable variable, $\pi_k$ is the assignment probability for each model, with $\sum_{k=1}^{K}\pi_k=1, (0<\pi_k<1)$, and $\theta_k$ is the internal parameters of the base distribution. 

Let $z_n$ be the latent variables for indicating categories.  

\begin{equation}
z_n \sim Categorical(z_n | \pi) \quad \sum_{k=1}^{K}z_{nk}=1
\end{equation}

where, $z_n=[z_{n1}, z_{n2}, ..., z_{nk}, ..., z_{nK}]$, in which only one element $z_{nk} = 1$. It means $x_n$ is correspondent to $\theta_k$.

If the base distribution is a Gaussian, then:

\begin{equation}\label{Equ: Gaussian}
P(x | \theta_k)= N(x | \mu_k,\Lambda^{-1}_k)
\end{equation}

where, $\mu_k$ is the mean vector and $\Lambda_k$ is the precision matrix. 

Therefore, an observable sample $x_n$ is drawn from GMM according to:
\begin{equation}\
x_n \sim \prod_{k=1}^{K}{{N(x_n | \nu_k,\Lambda_k)}^{z_{nk}}}
\end{equation}

As it is illustrated above, one crucial issue of GMM is to pre-define the number of components $K$. This is tricky because the probability distribution for each day's mobility is not identical. Thus, to define a fixed $K$ for all mobility GMM models is not suitable in our case.    

\subsubsection{Infinite Gaussian Mixture Model}

\begin{figure}[!t]
\centering
\includegraphics[width= \linewidth]{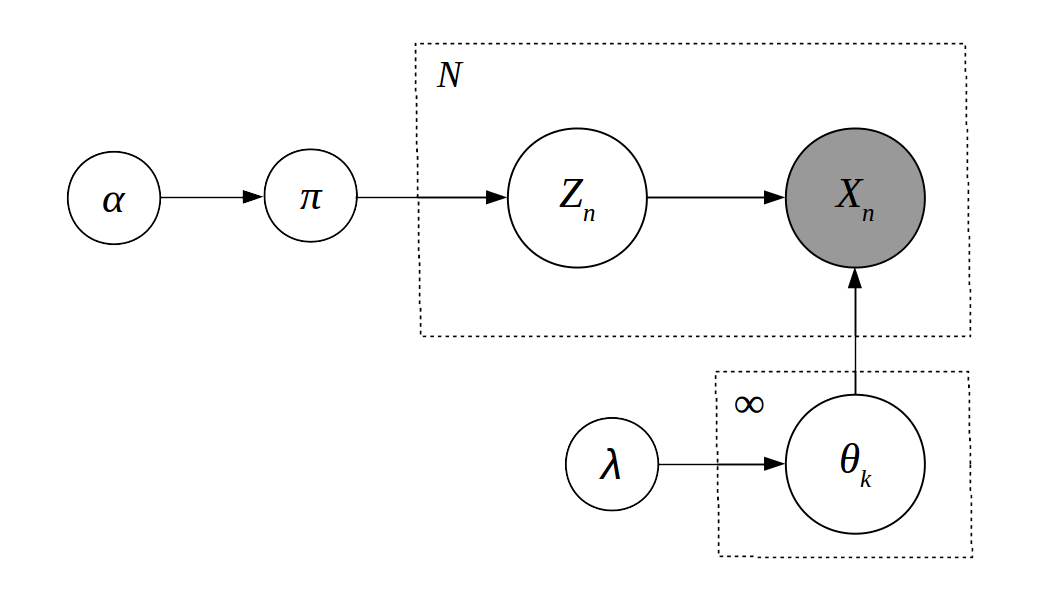}
\caption{The plate representation of Infinite Gaussian Mixture Model.}\label{Fig:IGMM_Plate}
\end{figure}

Alternatively, we resort to the Infinite Gaussian Mixture Model (IGMM) \cite{rasmussen2000infinite}. As compared to finite Gaussian Mixture Model, by using a Dirichlet process (DP) prior, IGMM does not need to specify the number of components in advance. Fig.~\ref{Fig:IGMM_Plate} presents the graphical structure of the Infinite Gaussian Mixture Model. 

In Fig.~\ref{Fig:IGMM_Plate}, the nodes represents the random variables and especially, the shaded node is observable and the unshaded nodes are unobservable. The edges represent the conditional dependencies between variables. And the variables are within the plates means that they are drawn repeatedly.

According to Fig.~\ref{Fig:IGMM_Plate}, the Dirichlet process can be depicted as:
\begin{equation}\label{Equ: Random Measure}
G \sim DP(\alpha, G_0)    
\end{equation}

where, $G$ is a random measure, which consists of infinite base measure $G_0$ and $\lambda$ is the hyper-parameter of $G_0$. In our case, it is a series of Gaussian distributions. And $\alpha \sim Gamma(1,1)$ is the concentration parameter. $N$ is the total samples number. $\theta_k$ is the parameters of base distribution. $X_k$ is the observable data for $\theta_k$. $Z_k$ is the latent variables that indicates the category of $X_k$. 

Alternatively, $G$ can be explicitly depicted as follow:

\begin{equation}\label{Equ: Dirichlet Process}
G(\theta) = \sum_{k=1}^{\infty}\pi_k \delta_{\theta_k}     
\end{equation}

where, $\theta_k \sim G_0(\lambda)$, and $\delta$ is Dirac function. $\pi_k$ determines the proportion weights of the clusters and the $\delta_{\theta_k}$ is the prior of the $\theta_k$ to determine the location of clusters in space.  

We choose the Stick-breaking process (SBP) \cite{sethuraman1994constructive} to implement the Dirichlet process as the prior for $\pi_k$.  
The the Stick-breaking process can be described as follow:

\begin{equation}\label{Equ: SBP Process}
\pi_k = \nu_k\prod_{j=1}^{k-1}{(1-\nu_{j})} \quad k\geq2
\end{equation}

where, $\nu_k \sim Beta(1,\alpha)$.

Since $P(x|\theta)$ is Gaussian, $\theta =\{\mu,\Lambda\}$. Further, let $G_0$ be a Gaussian-Wishart distribution, then, $\mu_k, \Lambda_k \sim G_0(\mu, \Lambda)$. Therefore, similarly, draw an observable sample $x_n$ from IGMM:
\begin{equation}\
x_n \sim \prod_{k=1}^{\infty}{{N(x_n|\nu_k,\Lambda^{-1}_k)}^{z_{nk}}}
\end{equation}

Then, Variational Inference is used to solve the IGMM models. As compared to Gibbs sampling or to a Markov chain Monte Carlo (MCMC) method which consumes a large mount of calculating time, Variational Inference is relatively fast \cite{blei2006variational}. The results will be demonstrated in the later experiments.         

 
\subsection{Measure Daily Trajectories Similarities}

The Kullback-Leibler (KL) divergence is a metric to evaluate the closeness between two distributions. For continuous variables, the KL divergence $D_{KL}(p||q)$ the expectation of the logarithmic difference between the $p$ and $q$ with respect to probability $p$ and vice versa. From (\ref{Equ: KLPQ}) and (\ref{Equ: KLQP}), it can be seen that the KL divergence is non-negative and asymmetric. In many occasions, the inequality of the KL divergence is notorious. However, in our methodology, on the contrary, we take advantage of the characteristics of inequality to reveal the similarities among different trajectories instead of the Jensen-Shannon divergence which is a symmetric metrics.

\begin{equation}\label{Equ: KLPQ}
D_{KL}(p||q) = \int_{-\infty}^{\infty} p(x)log(\frac{p(x)}{q(x)}) dx
\end{equation}

\begin{equation}\label{Equ: KLQP}
D_{KL}(q||p) = \int_{-\infty}^{\infty} q(y)log(\frac{q(y)}{p(y)}) dy
\end{equation}

There is no closed form to implement the KL divergence by the definition of (\ref{Equ: KLPQ}) and (\ref{Equ: KLQP}) for Gaussian Mixture Models. Instead, we resort to the Monte Carlo simulation method proposed in \cite{hershey2007approximating}. Then, the KL divergence can be caculated by:  

\begin{equation}\label{Equ: KLPQMC}
D_{KL_{MC}}(p||q) =\frac{1}{n} \sum_{i=1}^{n} log(\frac{p(x_i)}{q(x_i)})
\end{equation}

\begin{equation}\label{Equ: KLQPMC}
D_{KL_{MC}}(q||p) =\frac{1}{n} \sum_{i=1}^{n} log(\frac{q(y_i)}{p(y_i)})
\end{equation}

This method is to draw a large amount of i.i.d samples $x_i$ from distribution $p$ to calculate $D_{KL_{MC}}(p||q)$ according to (\ref{Equ: KLPQMC}) and $D_{KL_{MC}}(p||q) \to D_{KL}(p||q)$ as $n \to \infty$. And it is the same for implementing (\ref{Equ: KLQP}) by using (\ref{Equ: KLQPMC}). The results will be demonstrated in the later experiments. Furthermore, if we define a representative trajectory for a mobility pattern then we can distinguish whether a new trajectory belong to this cluster by comparing it to the representative trajectory. To do so, we need to set a threshold with a lower bound and an upper bound for the KL divergence, then it can be used as the metrics to cluster mobility patterns.


\subsection{Discover Mobility Patterns}

As mentioned before, our task is to find the trajectories which are mutually similar. For this reason, we treat the different mobility patterns as different clusters in which the daily trajectories are their sub-members. Even so, the trajectories within the same clusters still can not be treated as identically distributed as other conventional clustering methods because of different trajectory lengths. Hence, we need to devise a algorithm that is able to cluster the trajectories based on the distribution similarity and the aforementioned KL divergence can be applicable as closeness metrics. Note that due to the large data scale and the number of the potential clusters, a high accuracy solution is intractable sometimes. Therefore, instead of pursuing a very accurate result, our purpose is to reach a relative accurate result in a reasonable amount of calculating time.   

The algorithm we devise is shown in Algorithm~\ref{Alg:Algorithm} and its variables are described in Table~\ref{Table:Variables}.

The first step of the clustering algorithm is to calculate the probability densities using the Infinite Gaussian Mixture Models. At this step, we create a list, in which the members are the probability densities of each. Then, the first cluster is created with one trajectory as its first member and it also will be compared with other trajectories.

Afterwards, we select another daily trajectory in the list and calculate the KL divergences, both $D_{KL(p||q)}$ and $D_{KL(q||p)}$. And the new trajectory is added to the current cluster if the minimum and maximum of the KL-divergences are smaller than the lower bound and upper bound of the thresholds at the same time, respectively. And if the $D_{KL(p||q)}$ is smaller than  $D_{KL(q||p)}$, the new trajectory become the benchmark for the current cluster. An alternative way to do this is to compute the probability density of the current cluster using all the data of the discovered trajectories, however, the calculation will be massive.

This step will be repeated until all the trajectories belonging to the current cluster are discovered at the end of this iteration. Then, all the members of the current cluster are removed from iteration because, we assume that each trajectories can only be a member of one mobility pattern. At the start of new iteration, a new cluster is created, repeat the above steps until the list is empty. Finally, all the mobility patterns are discovered.    


\captionsetup{font={footnotesize,sc},justification=centering,labelsep=period}%
\begin{table*}[htbp]
\caption{Variables Description}\label{Table:Variables}
\centering%
\begin{tabular}{cll}
\hline
\textit{Variable} & \textit{Domain} & \textit{Description} \\
\hline

$d$ & $\{1, 2, \dots, D\}$ & Number of data collecting day\\
$X$ & $\{X_1, X_2, \dots, X_d, \dots,  X_D\}$ & Total GPS data (longitudes, latitudes \\ 
$P$ & $\{P_1, P_2, \dots, P_d, \dots, P_D\}$ & Probability density for $X$ \\
$M$ & $\{M_1, M_2, \dots, M_k, \dots M_K\}$ & Total mobility patterns \\  
$M_k$ & $\{X_{k1}, X_{k2}, \dots, X_{kn}\}$ & Discovered mobility pattern \\ 
$Th$ & $\{lower bound, upper bound\}$  & Threshold for distinguishing patterns\\
$D_{KL}$ & $\{D_{KL(p||q)}, D_{KL(q||p)}\}$& KL divergences \\

\hline
\end{tabular}
\end{table*}
\captionsetup{font={footnotesize,rm},justification=centering,labelsep=period}%


\begin{algorithm} 
\caption{Mobility Pattern Discovering Algorithm}
\label{Alg:Algorithm}
\begin{algorithmic}[1]
\Require{$X$} 
\Ensure{$M$}
\Statex
\State{$P$ $\gets$ IGMM($X$)} \Comment{probability density estimation}
\Statex
\State{Initialize:$M= \{M_k\}$} \Comment{create the mobility patterns set}
\While{$P\neq\emptyset$}
    
    \State{$X_s = X_1$} \Comment{set the baseline mobility for $M_k$}
    \State{$M_k= \{X_s\}$} \Comment{create current pattern $M_k$}
    \For{$d=2, \dots, D$}                    
        \State {$D_{KL}$ $\gets$ {$(P_s,P_d)$}}\Comment{measure similarity}
        \If{($\min(D_{KL})<Th[0]$) \& ($\max(D_{KL})<Th[1]$)}\Comment{two patterns are similar}
            \State{add $P_d$ to $M_k$}\Comment{add new member}
            \If{$D_{KL}[0]>D_{KL}[1]$}
                \State{$P_s$ $\gets$ $P_d$} \Comment{change the baseline mobility}
            \EndIf 
        \EndIf
    \EndFor
    \State{remove $P_d \in M_k $ from $P$}\Comment{current pattern is finished}
\Statex    
    \State{create $M_{k+1}$} \Comment{find new mobility pattern}
    \State add $M_{k+1}$ to $M$ 
\EndWhile\\
\Statex
\Return{$M$}
\end{algorithmic}
\end{algorithm}

As it can be seen that our algorithm is designed to discover the latent mobility patterns automatically without the pre-knowledge of the numbers of existing patterns.

 \section{Experiments and Results} \label{Sec: Experiments and Results}
 
 \subsection{Dataset Description}
 
 We use the Mobile Data Challenge (MDC) dataset \cite{kiukkonen2010towards}, \cite{laurila2012mobile} to validate our method. This dataset records comprehensive smartphone usages with fine granularity of time. The participants of the MDC dataset are up to nearly $200$ and the data collection campaign lasts more than $18$ months. This abundant information thus can be used to investigate individual mobility patterns.

 To collect the individual location information, as compared to other methods, for instance, through stand-alone GPS devices, using GPS-equipped smartphones is a more practical way to have a larger group of participants without affecting their daily life.

 In our study, we attempt to find the trajectories that belong to the same mobility patterns, thus we focus the spatial information of the GPS records, namely, the latitudes and longitudes and the time-stamps of the data are not considered. Meanwhile, since we consider not only the significant places but all location records, we use the unlabeled data without any semantic information.       
 
\subsection{Experimental Setup}
 
\begin{figure}[!t]
    \centering
    \includegraphics[width= \linewidth]{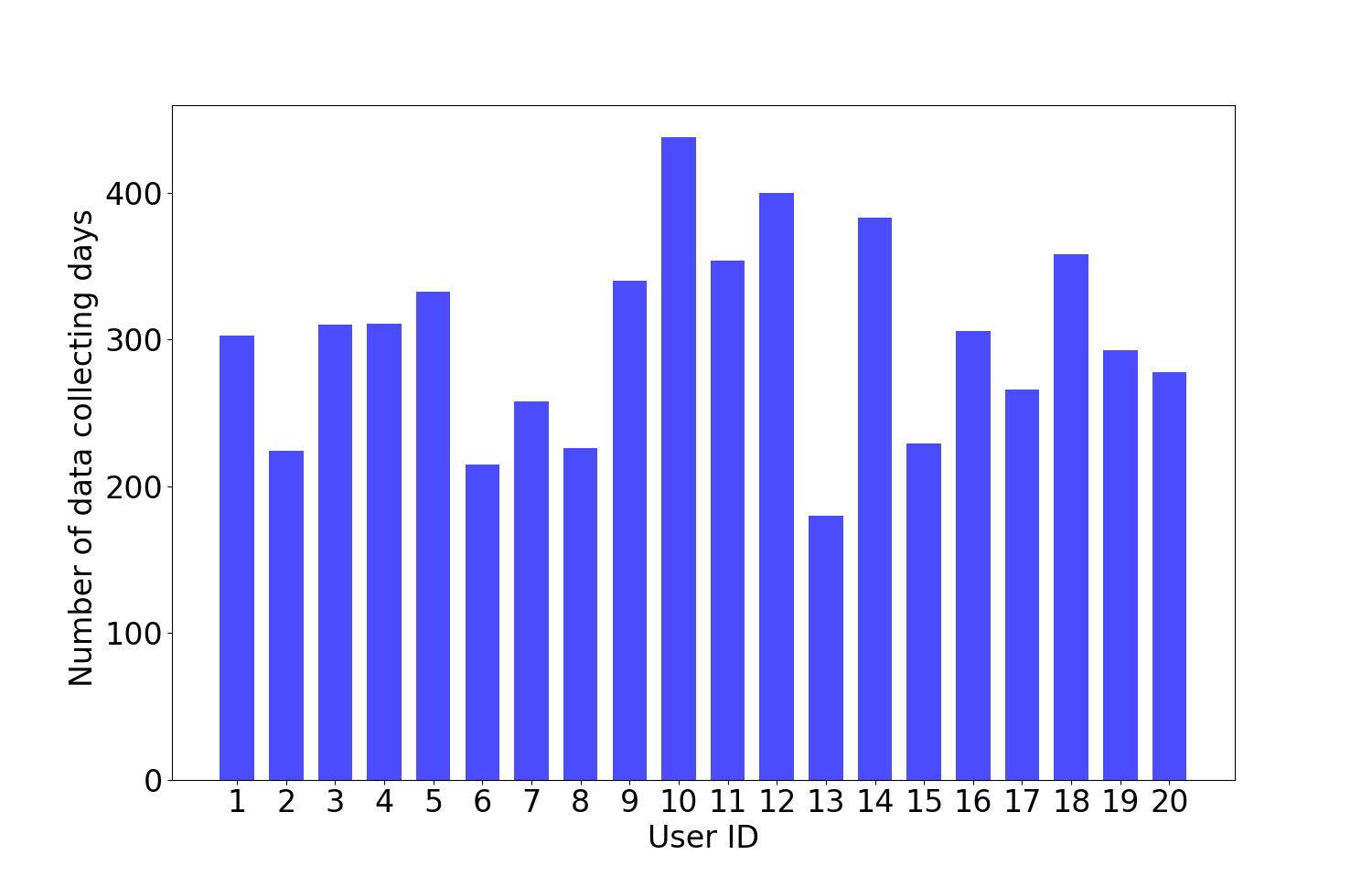}
    \caption{Number of data collecting days for each user}\label{Fig: DayNum}
\end{figure}

In the conducted experiment, we randomly select $20$ users with sufficient data. And each user's is segmented by the time range of one day. Fig.~\ref{Fig: DayNum} demonstrates the number of data collecting days for each user. It can be seen that the data collecting days for most users are more than $200$. And with such amount of data, we believe that it is possible to discover individual's mobility patterns from it.       

\begin{figure}[!t]
    \centering
    \includegraphics[width= 0.9\linewidth]{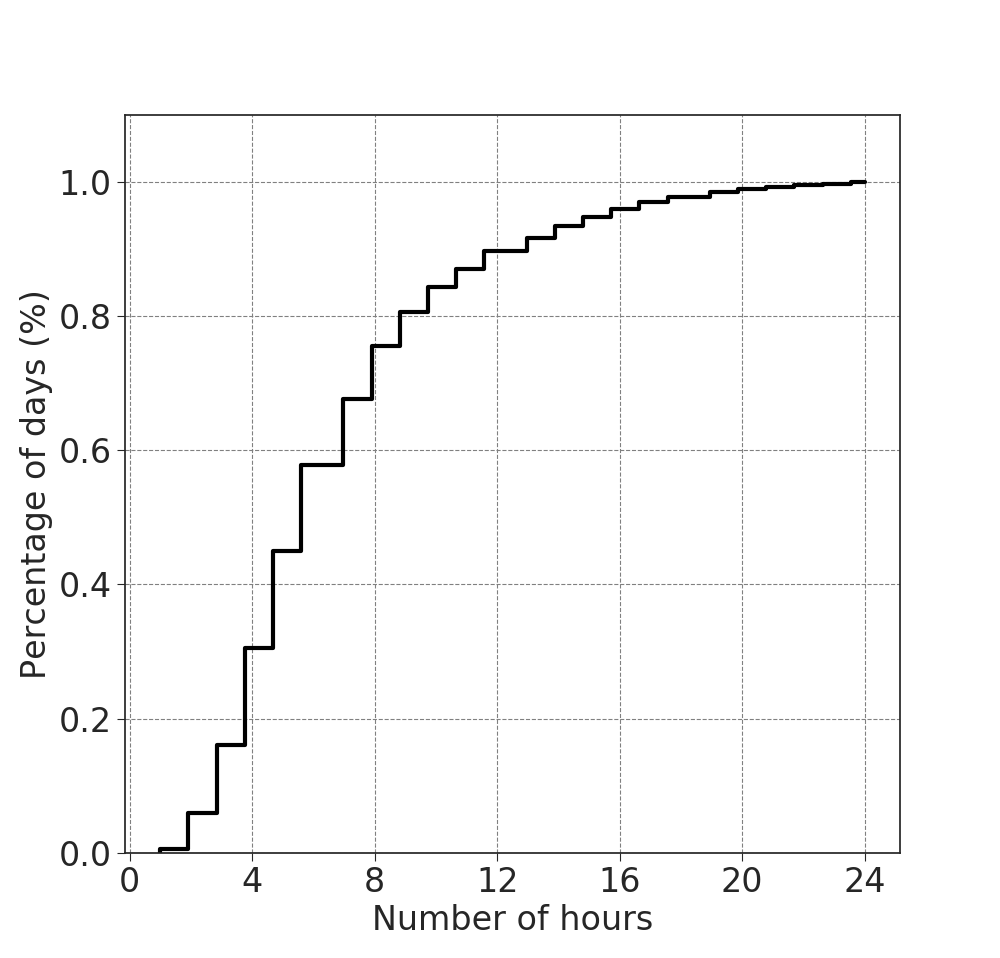}
    \caption{Empirical cumulative distribution of hours per data collecting days.}\label{Fig: HourCDF}
\end{figure}

However, as it is illustrated in Fig.~\ref{Fig: HourCDF}, the data length of each day varies from less than $4$ hours to $24$ hours. And most of them is less than $8$ hours. Hence, we also should be aware that some data can be missing because the GPS modules were turned off or were not functioning. Consequently, it is one of the reasons that cause the data sparsity problem. In the following part, we will prove that our method can mitigate the impact of data sparsity.

Table \ref{Table: DataDescription} summarizes the temporal information about the GPS data for conducting the experiments.

\captionsetup{font={footnotesize,sc},justification=centering,labelsep=period}%
\begin{table}[htbp]
\caption{Data collecting time }\label{Table: DataDescription}
\centering%
\begin{tabular}{lcc}
\hline
 &\textit{Average} & \textit{Total}\\
\hline
Collecting days for all users & 300.25 days & 6005.0 days \\
Collecting hours per day for all users & 6.93 hours& 41595.0 hours\\
Collecting hours per day for each user & 6.67 hours & 2084.65 hours \\

\hline
\end{tabular}
\end{table}
\captionsetup{font={footnotesize,rm},justification=centering,labelsep=period}%

To test the performance of our method, we will conduct three experiments from different perspectives:

\begin{itemize}
    \item We compare the IGMM model with the GMM model on estimating the daily trajectories probability density.
    \item We use that the KL divergence to measure the closeness of different trajectories. 
    \item We test our method on each selected user data so as to find the daily mobility patterns for each individual.  
    \item We compare the results of the IGMM models to a series of fixed-number components GMM models. 
    \item We run the algorithm on the varying-length datasets, in the aim to find the minimum data length for discovering most mobility patterns of one individual.    
    
\end{itemize}

\subsection{Experimental Results}

\subsubsection{Probability Density Estimation}

\begin{figure}[!t]
\centering
\includegraphics[width= \linewidth]{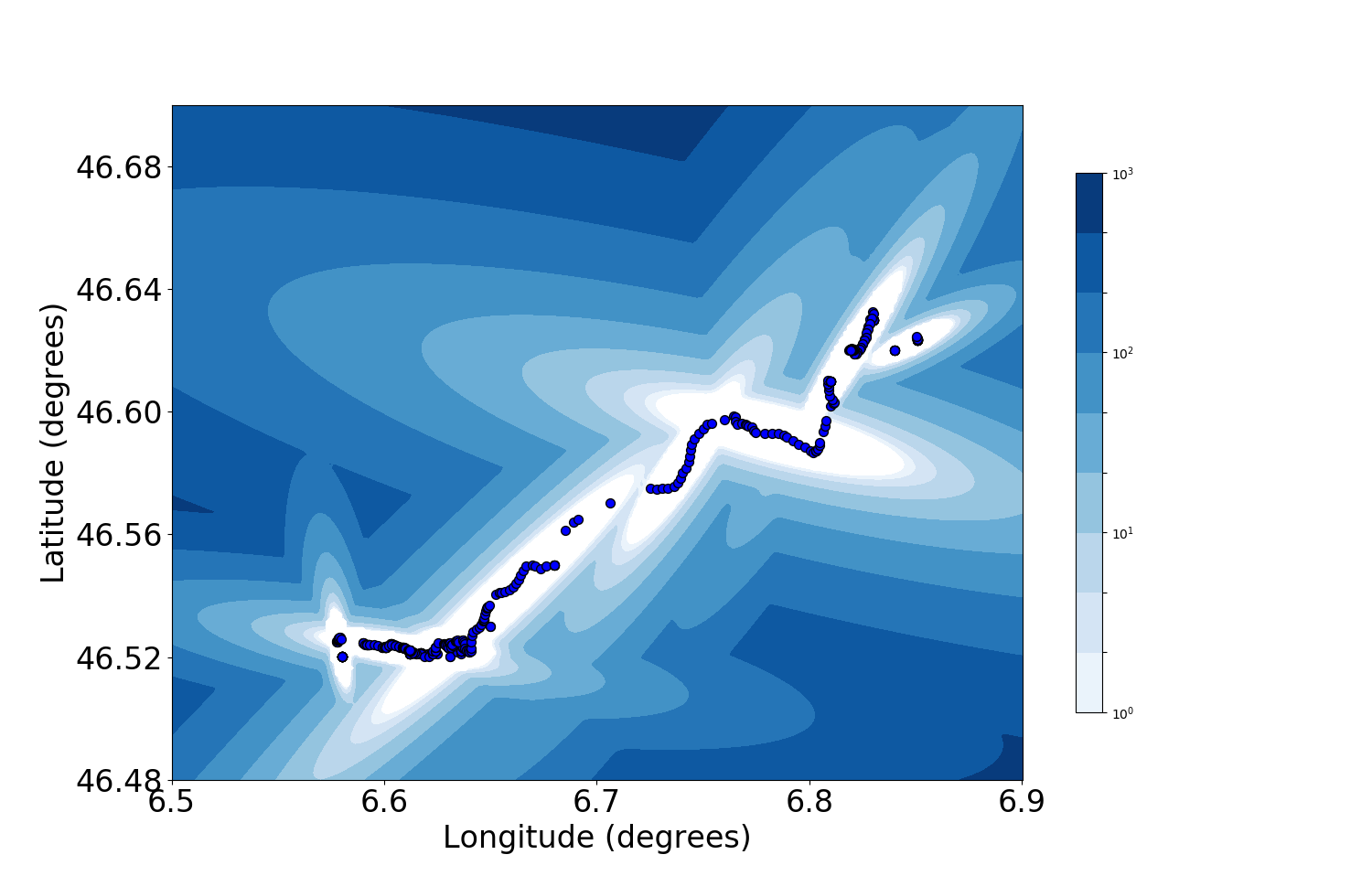}
\caption{Distribution estimation by GMM (negative log-likelihood)}\label{Fig:GMM}
\end{figure}

\begin{figure}[!t]
\centering
\includegraphics[width= \linewidth]{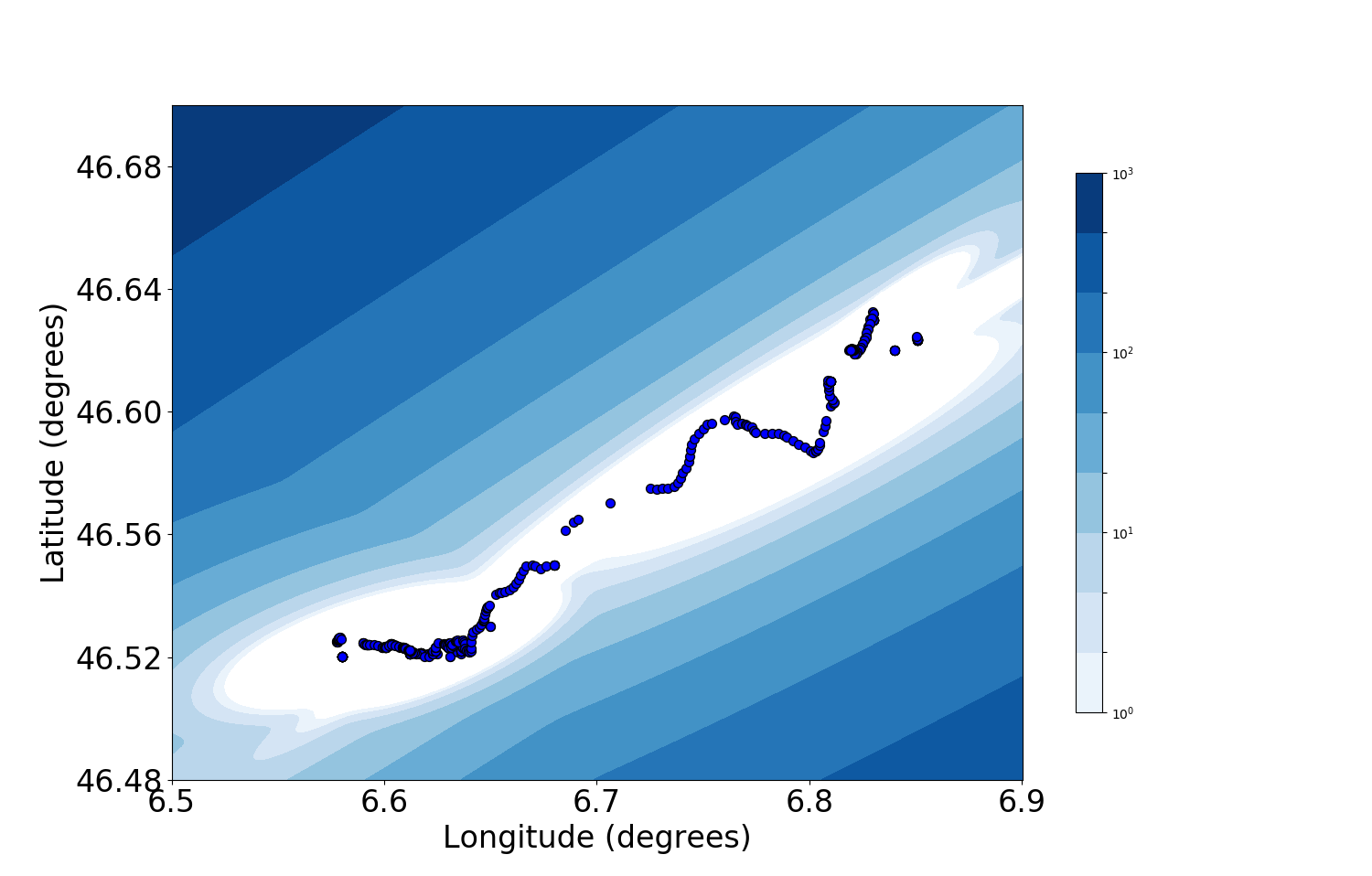}
\caption{Distribution estimation by IGMM (negative log-likelihood)}\label{Fig:IGMM}
\end{figure}

Fig.~\ref{Fig:GMM} and Fig.~\ref{Fig:IGMM} show the density estimation results obtained by GMM and IGMM, respectively. It can be seen that, compared to the GMM model, the result of the IGMM model is more smooth. It suggests that IGMM is not affected by the number of components and it infers more information from the original data and it is less influenced by data sparsity. That is to say, on the same dataset, the computational results of IGMM have higher fidelity. Hence, in our approach, we chose IGMM to estimate probability density of daily mobility. 

\subsubsection{Measuring Daily Trajectories Similarities}

\begin{figure*}[!t]
    \centering
    \begin{subfigure}[b]{0.4\linewidth}
        \includegraphics[width=\linewidth]{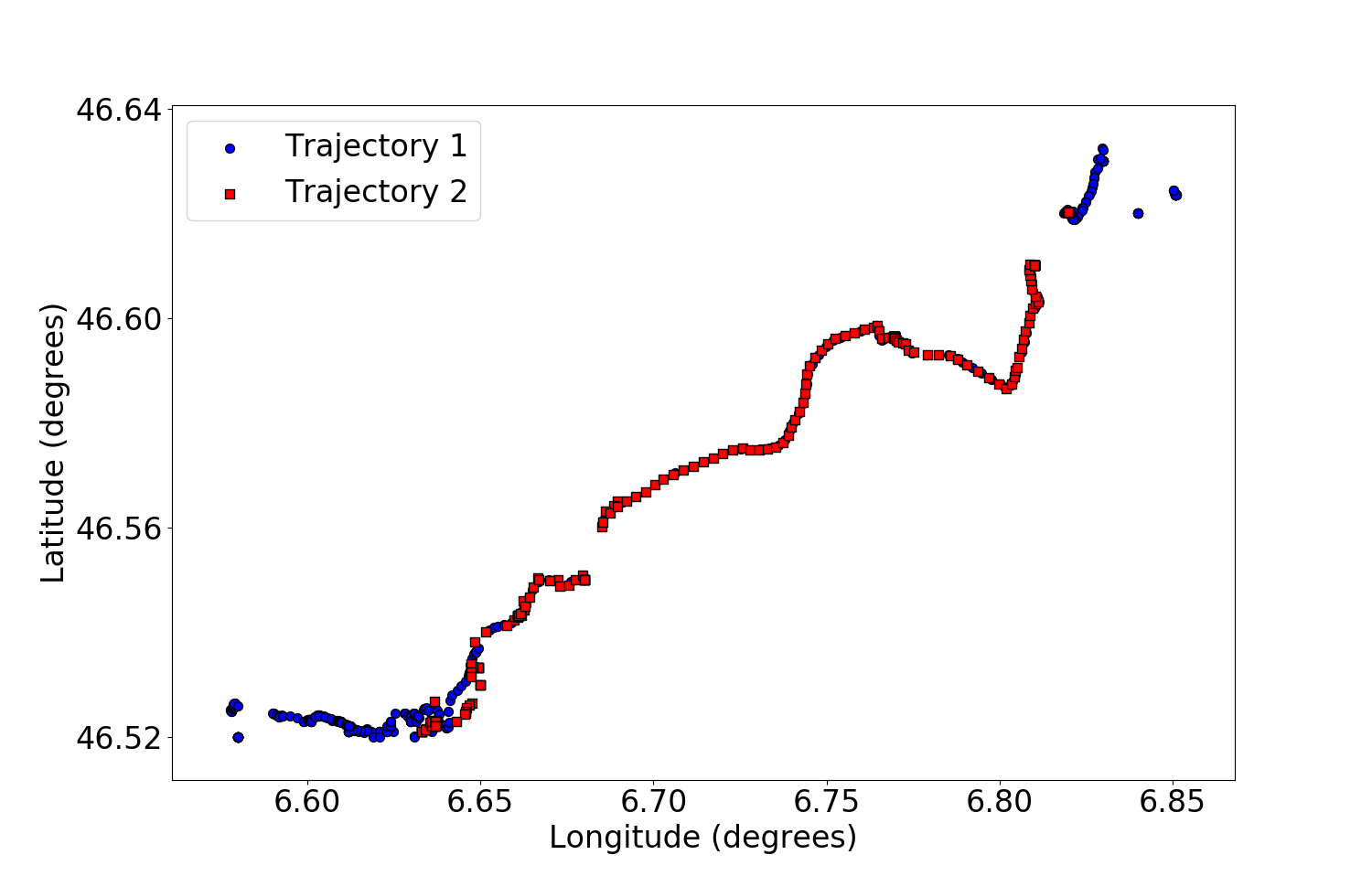}
        \caption{Trajectory 1 and Trajectory 2}\label{Fig: T1T2}
    \end{subfigure}
    ~ 
    \begin{subfigure}[b]{0.4\linewidth}
        \includegraphics[width=\linewidth]{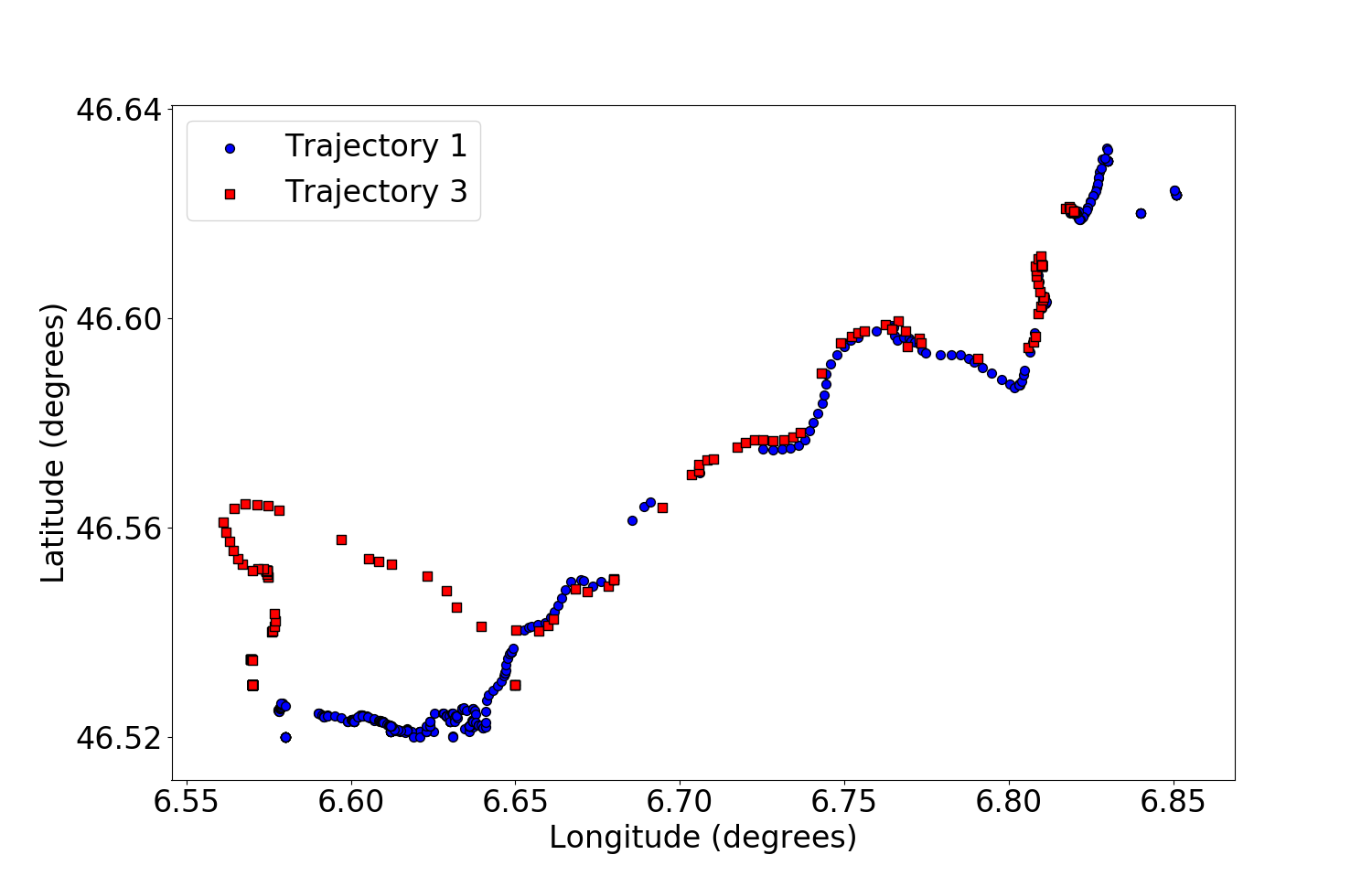}
        \caption{Trajectory 1 and Trajectory 3}\label{Fig: T1T3}
    \end{subfigure}

    \begin{subfigure}[b]{0.4\linewidth}
        \includegraphics[width=\linewidth]{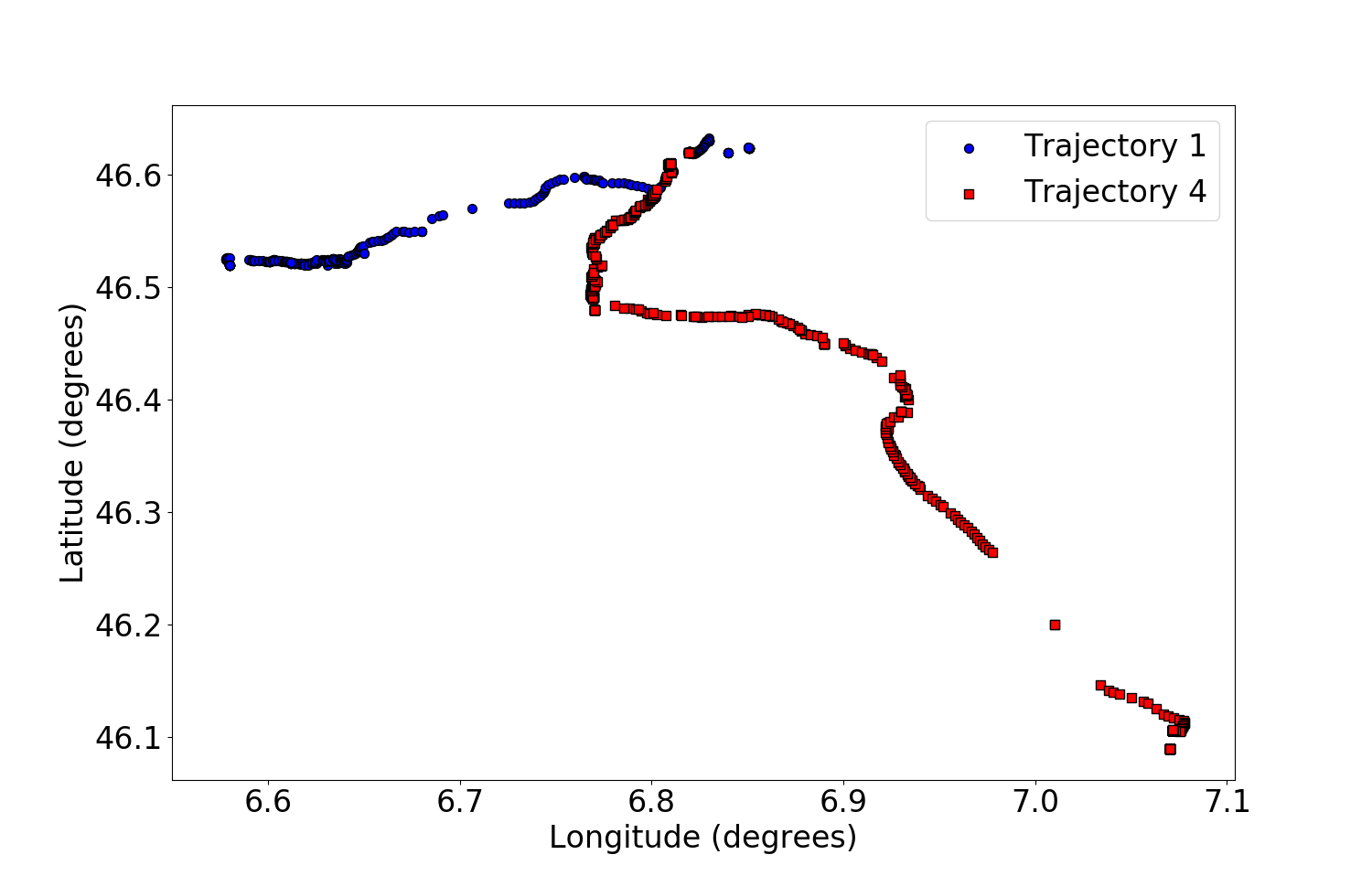}
        \caption{Trajectory 1 and Trajectory 4}\label{Fig: T1T4}
    \end{subfigure}
    ~ 
    \begin{subfigure}[b]{0.4\linewidth}
        \includegraphics[width=\linewidth]{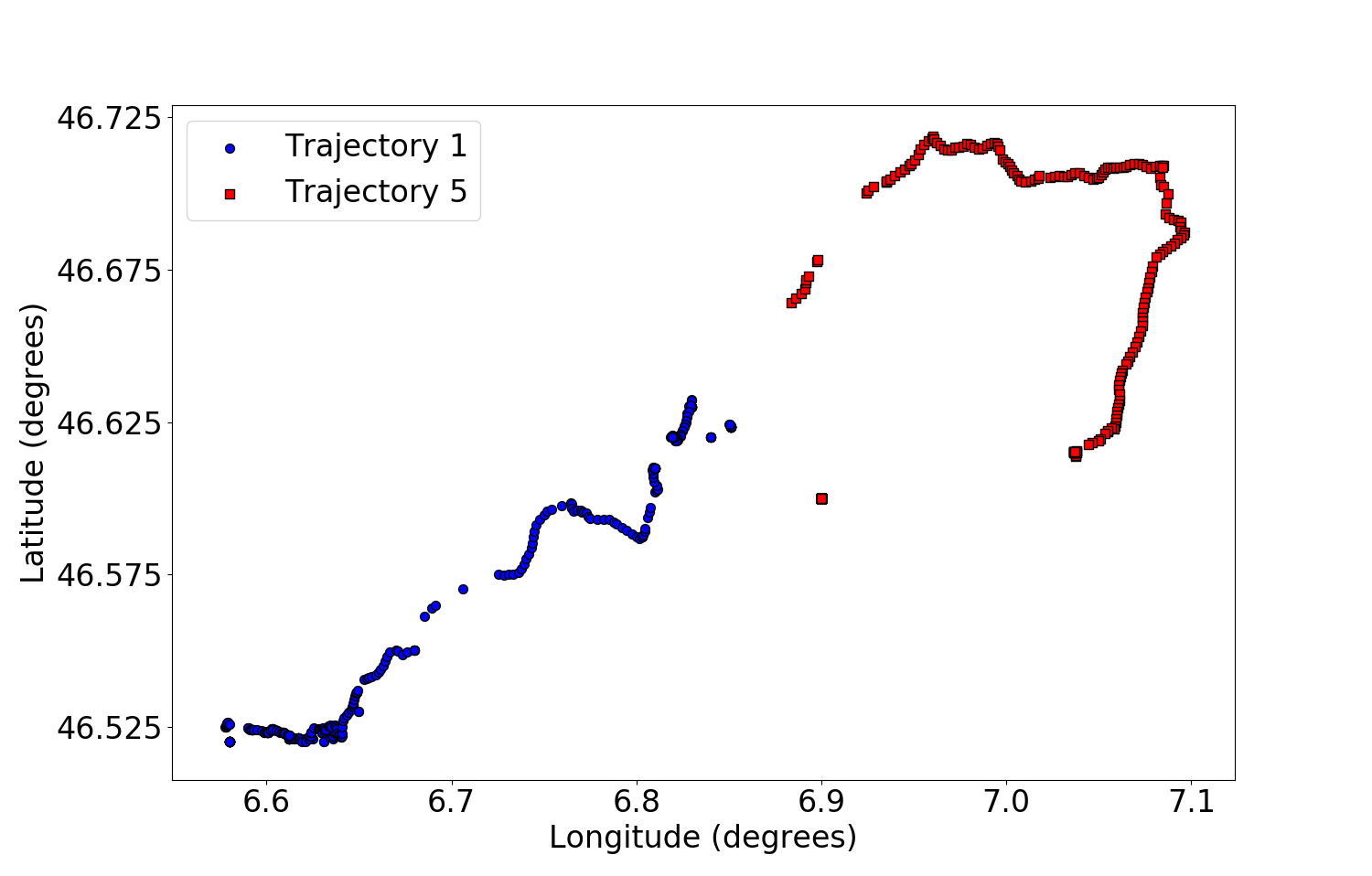}
        \caption{Trajectory 1 and Trajectory 5}\label{Fig: T1T5}
    \end{subfigure}

    \caption{Comparison Between Different Trajectories.}\label{Fig: Trajectories}
\end{figure*}

As shown in Fig.~\ref{Fig: Trajectories}, we select $5$ daily trajectories from the data of one random user to present the KD divergences between different trajectories. The baseline trajectory is the Trajectory 1 and the rest of trajectories are chosen to make comparisons.  

\captionsetup{font={footnotesize,sc},justification=centering,labelsep=period}%
\begin{table}[htbp]
\caption{KL-Divergences for Different Trajectories.}\label{Table: KL-Divergence}
\centering%
\begin{tabular}{cccc}
\hline
\textit{p} & \textit{q} & \textit{$D_{KL(p||q)}$} & \textit{$D_{KL(p||q)}$} \\
\hline

Trajectory 1 & Trajectory 2 & 7.21 & 2.82 \\
Trajectory 1 & Trajectory 3 & 1.28 & 1.83 \\
Trajectory 1 & Trajectory 4 & 19.07 & 1269.47 \\
Trajectory 1 & Trajectory 5 & 3.08 & 996.17 \\

\hline
\end{tabular}
\end{table}
\captionsetup{font={footnotesize,rm},justification=centering,labelsep=period}%

The combinations are shown in Figure~\ref{Fig: Trajectories} and the results are illustrated in Table~\ref{Table: KL-Divergence}:

Trajectory 2 is nearly a subset of Trajectory 1 and thus $D_{KL}(p||q)$ is larger than $D_{KL(p||q)}$. And their values are both small, thus Trajectory 2 and Trajectory 1 can be regarded to belong to the same mobility pattern. Trajectory 3 is very similar to Trajectory 1 and  $D_{KL}(p||q)$ almost equals to $D_{KL}(q||p)$. Hence, they also are the members of the same mobility pattern. Trajectory 4 share a small part with Trajectory 1 whereas generally they are very different. $D_{KL}(p||q)$ and $D_{KL}(q||p)$ are both very large. Therefore, it is reasonable to recognize Trajectory 4 and Trajectory 1 as different patterns. And Trajectory 5 is totally different from Trajectory 1. And $D_{KL}(p||q)$ is small but $D_{KL}(p||q)$ are very large. So they naturally are not in the same pattern. According to the trajectories in the Fig.~\ref{Fig: Trajectories} and the results in Table~\ref{Table: KL-Divergence}, it shows that the KL divergence is able to illustrate the difference among trajectories and can be the metrics for clustering. 

\subsubsection{Discovering Daily Mobility Patterns}
We run our algorithm on the data of the $20$ users to discover their daily mobility patterns.

\begin{itemize}
    \item \textbf{Discovered Patterns}: The partial results for different randomly selected user data are demonstrated in Fig.~\ref{Fig: Patterns}. It shows that, after clustered by our proposed algorithm, the data is split into different mobility patterns. Each cluster is composed of trajectories close to each other even if they are not distributed with the same density in the space. That proves our methodology is able to find the different mobility patterns even under the condition of noisy data and discontinuous trajectories.   


 \begin{figure}[!t]
    \centering

    \begin{subfigure}[b]{0.45\linewidth}
        \includegraphics[width=\linewidth]{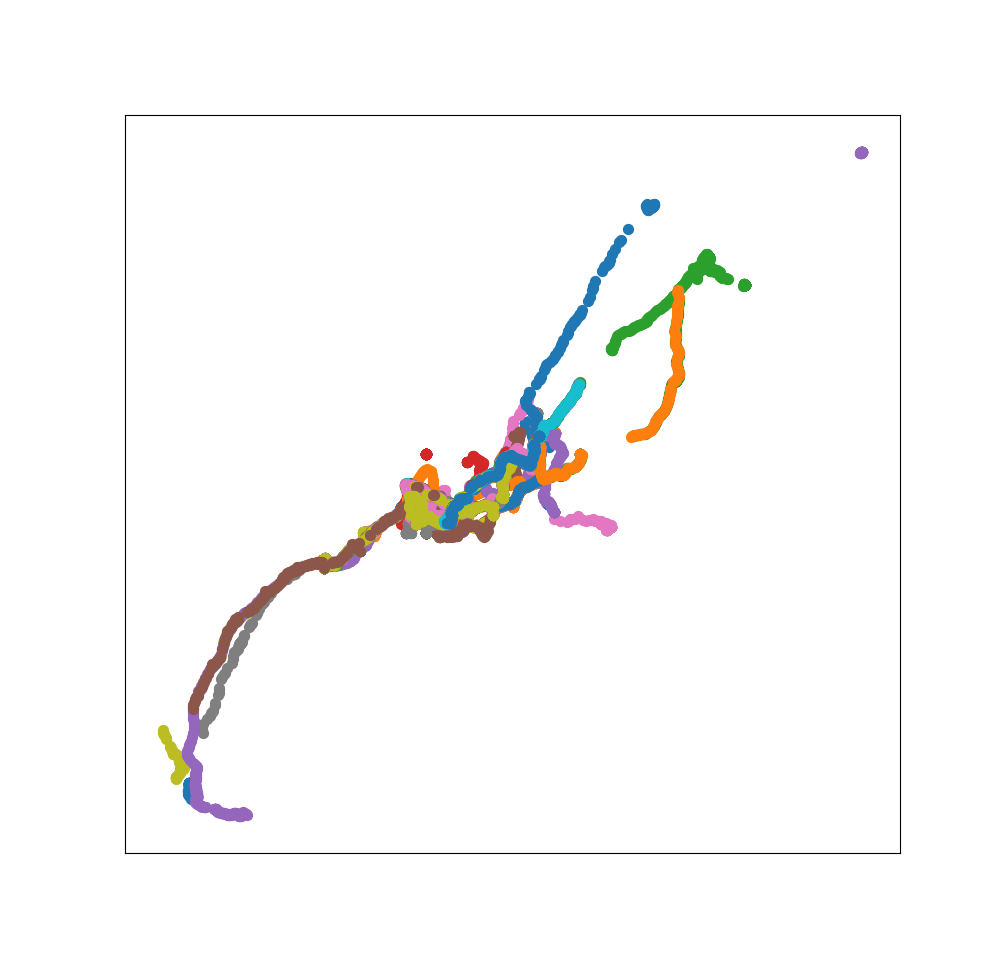}
    \end{subfigure}
    ~   
    \begin{subfigure}[b]{0.45\linewidth}
        \includegraphics[width=\linewidth]{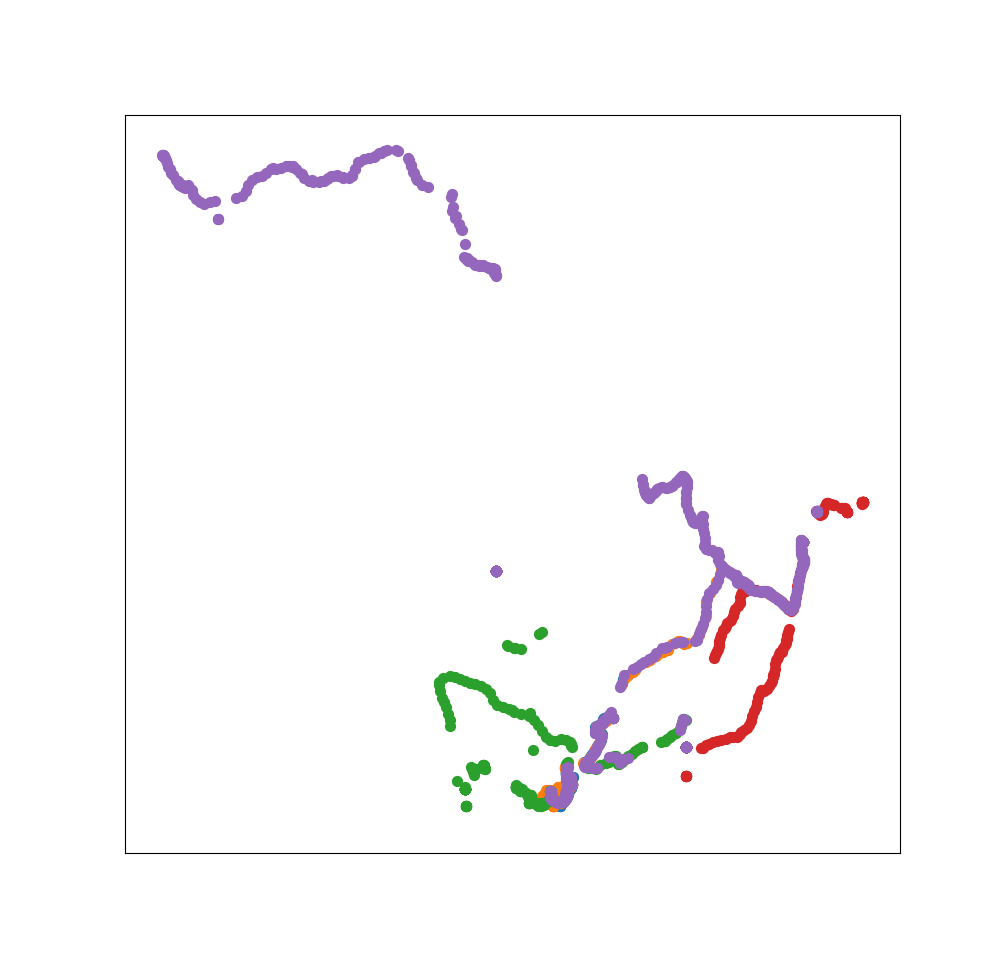}
    \end{subfigure}
    ~    
    \begin{subfigure}[b]{0.45\linewidth}
         \includegraphics[width=\linewidth]{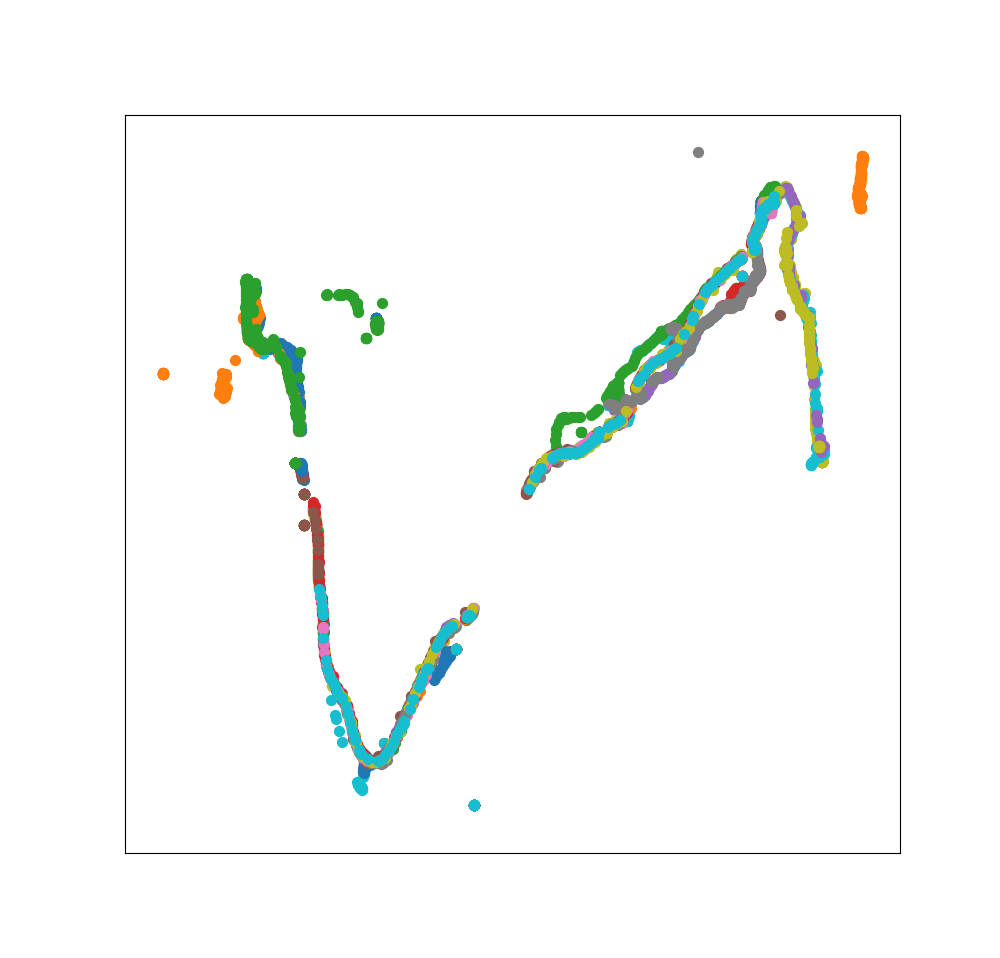}
    \end{subfigure}
    ~  
    \begin{subfigure}[b]{0.45\linewidth}
        \includegraphics[width=\linewidth]{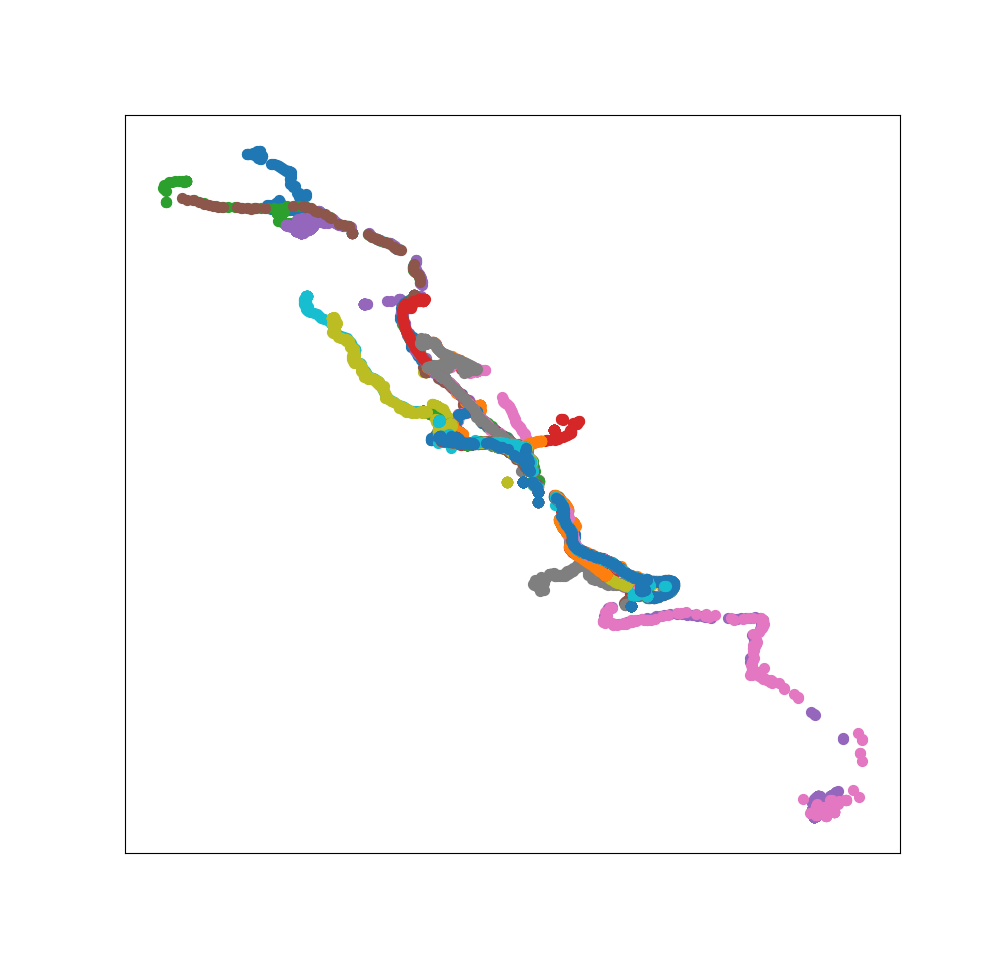}
    \end{subfigure}
    ~  
    \begin{subfigure}[b]{0.45\linewidth}
        \includegraphics[width=\linewidth]{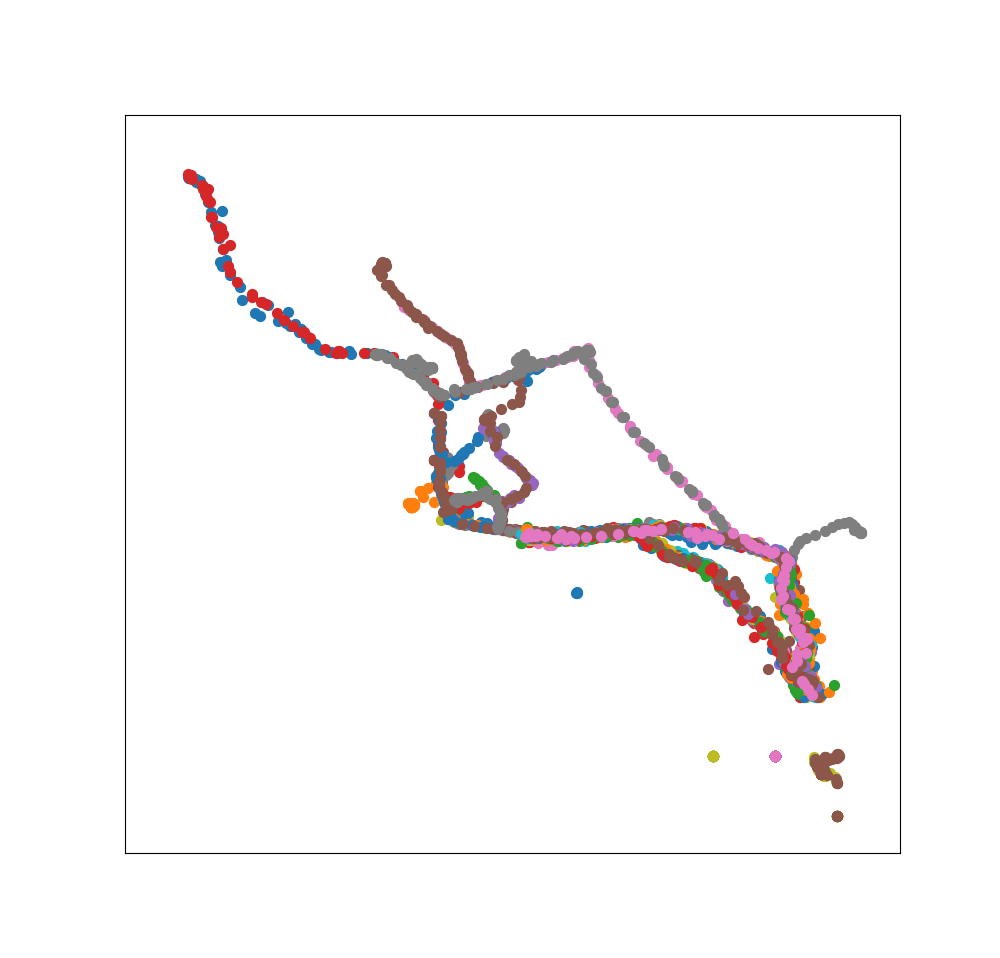}
    \end{subfigure}
    ~  
    \begin{subfigure}[b]{0.45\linewidth}
        \includegraphics[width=\linewidth]{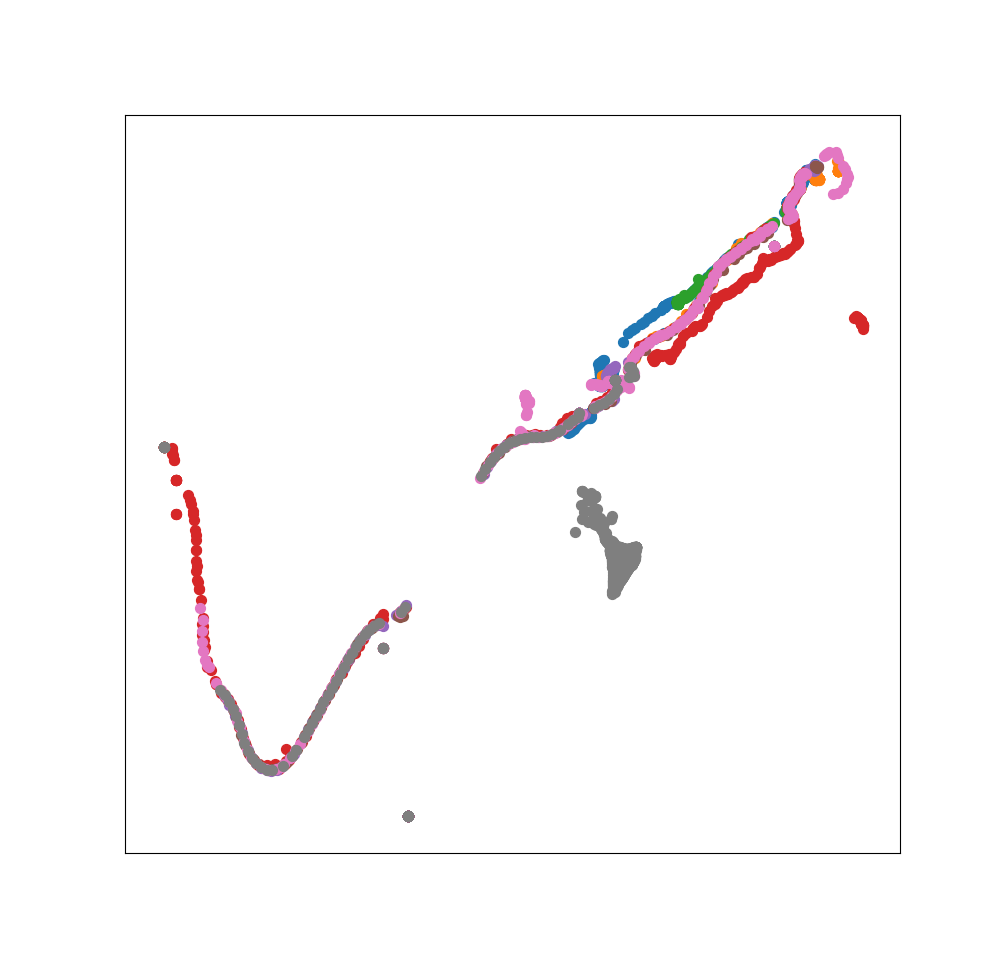}
    \end{subfigure}

    \caption{Discovered mobility patterns from three random selected users. Different colors represents different days.}\label{Fig: Patterns}
\end{figure}
 
Fig.~\ref{Fig: Representatives} shows that our methodology is not only able to identify the different patterns in the daily trajectories data but is also able to find the most representative trajectories for each mobility pattern. 
 
 \begin{figure}[!t]
    \centering

    \begin{subfigure}[b]{0.45\linewidth}
        \includegraphics[width=\linewidth]{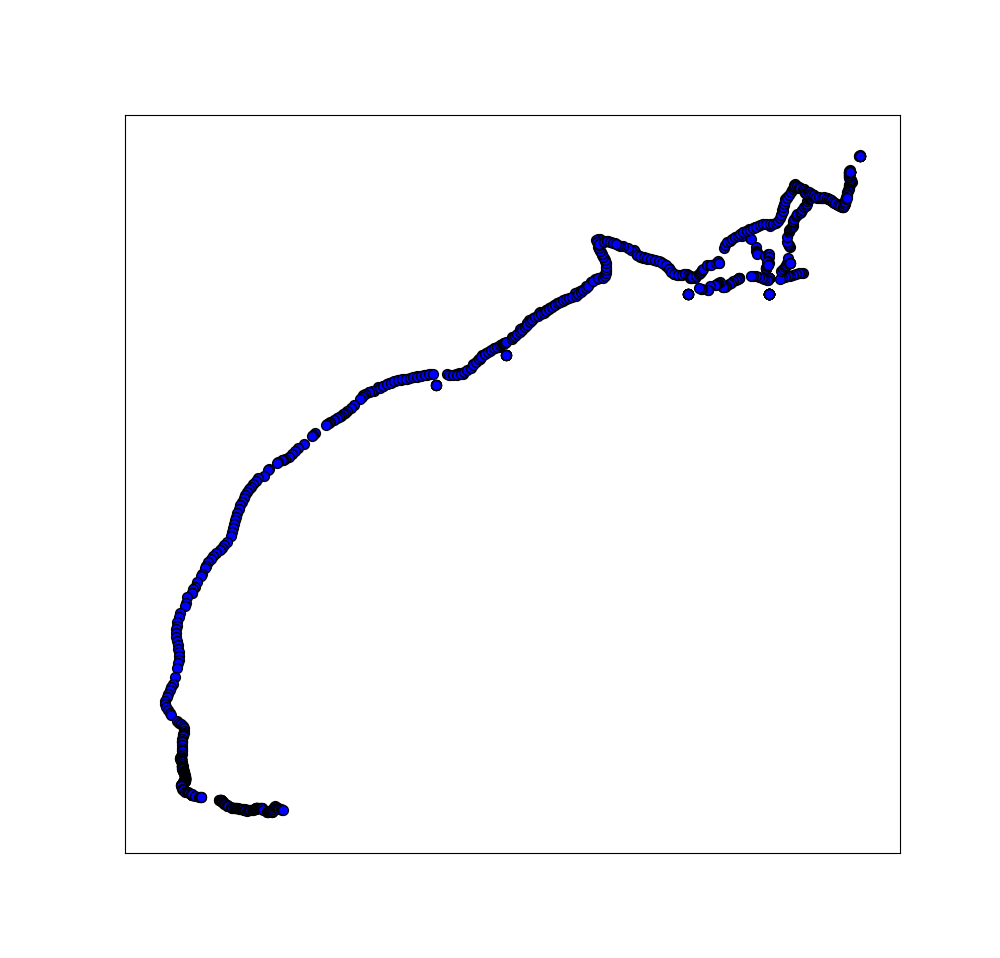}
    \end{subfigure}
    ~  
    \begin{subfigure}[b]{0.45\linewidth}
        \includegraphics[width=\linewidth]{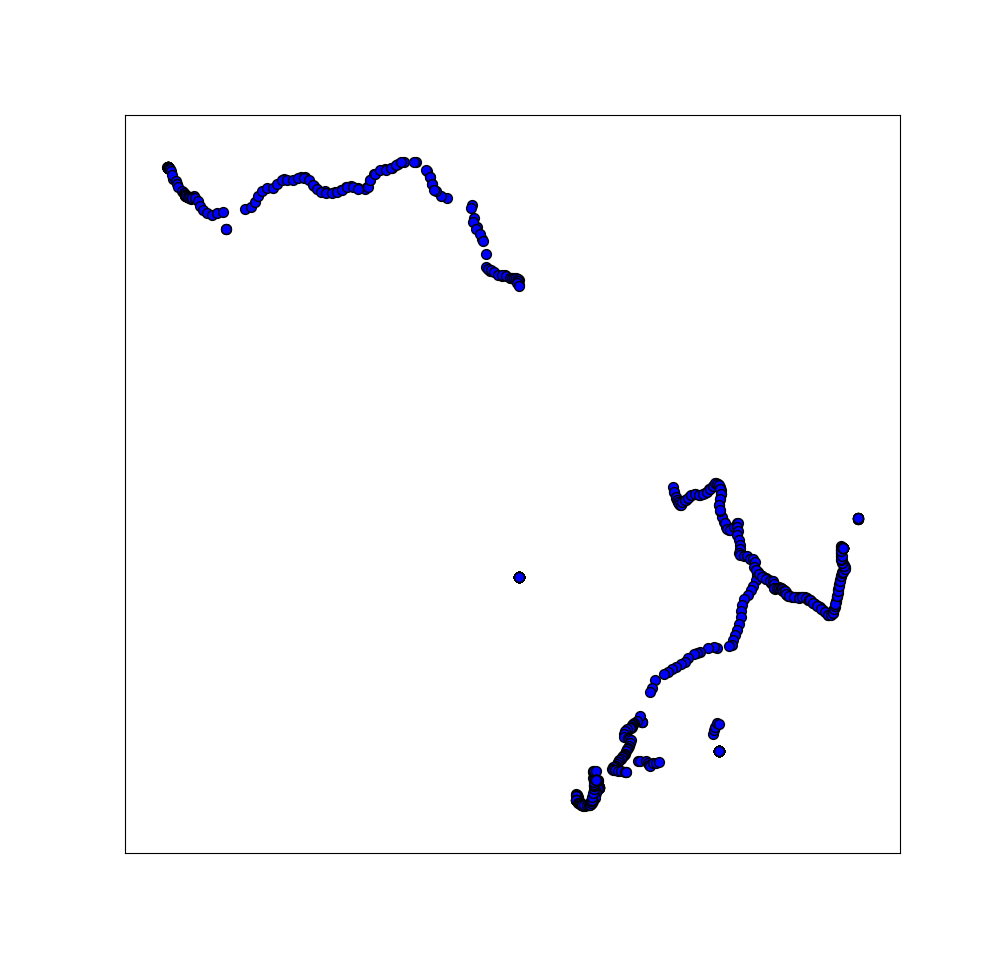}
    \end{subfigure}

    \begin{subfigure}[b]{0.45\linewidth}
         \includegraphics[width=\linewidth]{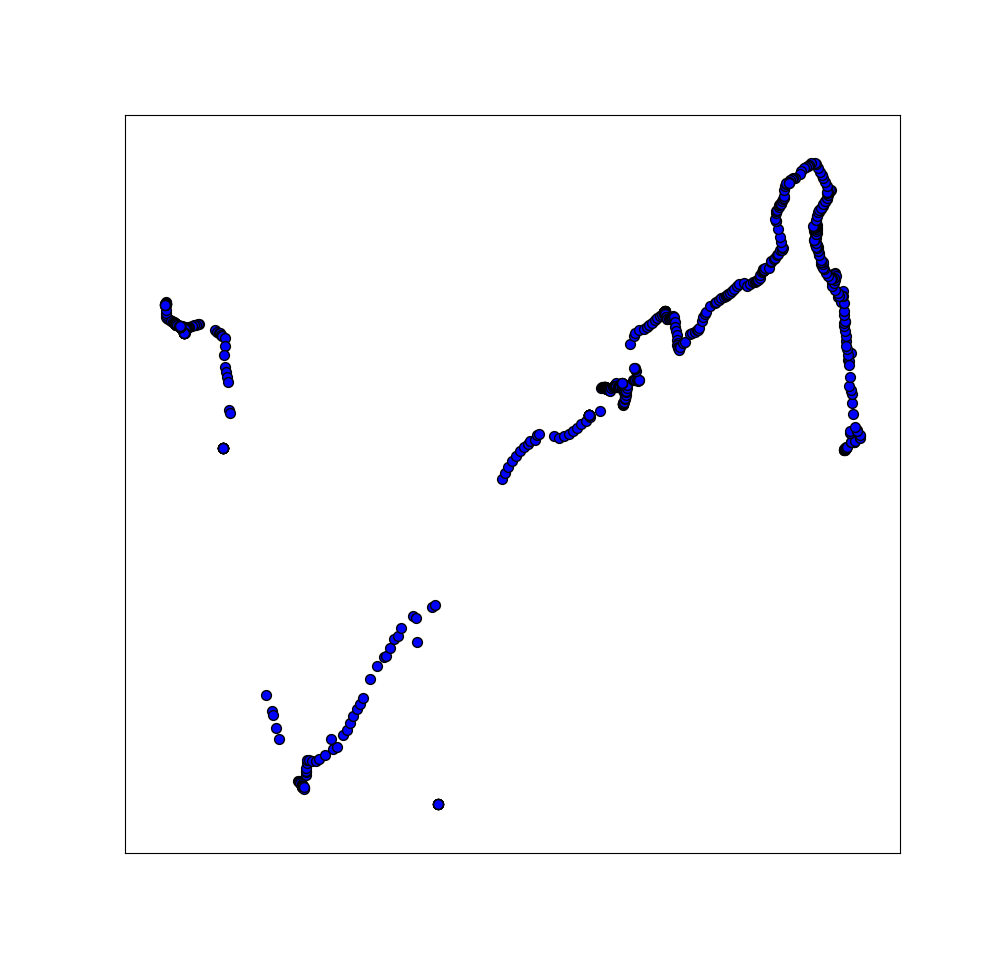}
    \end{subfigure}
    ~      
    \begin{subfigure}[b]{0.45\linewidth}
        \includegraphics[width=\linewidth]{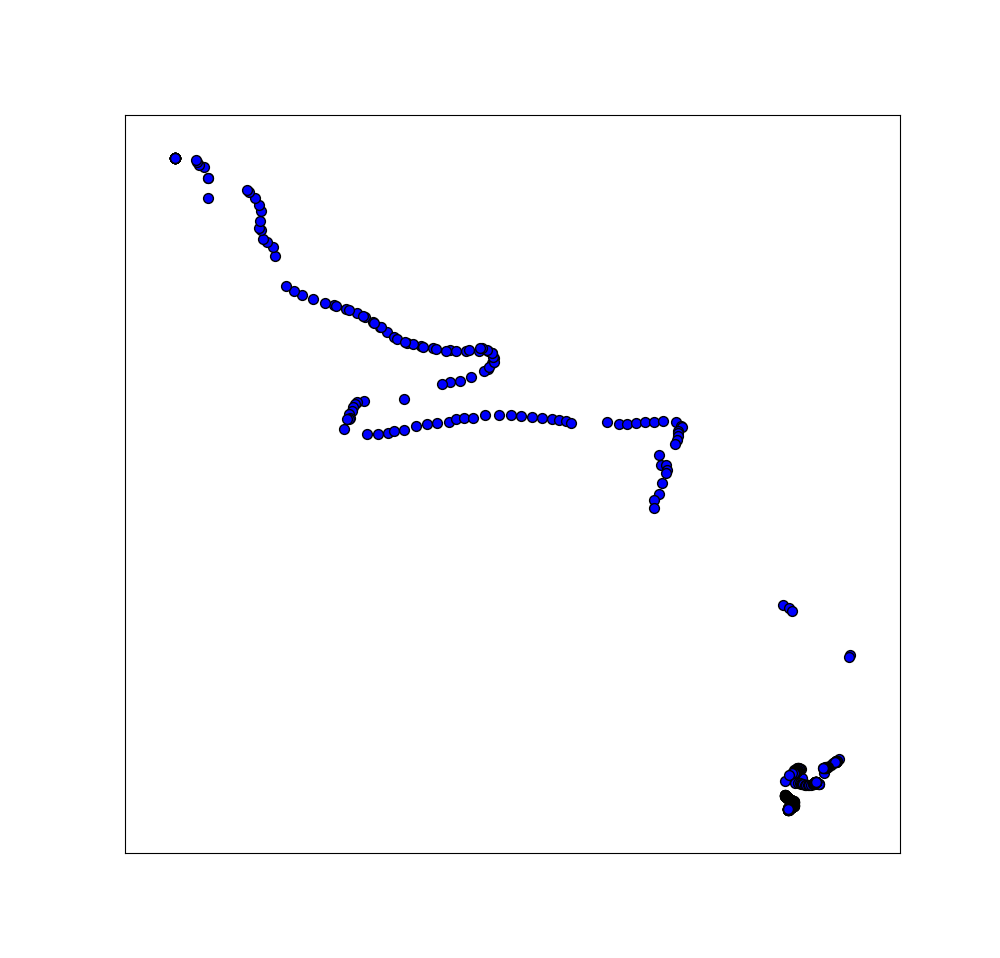}
    \end{subfigure}

    \begin{subfigure}[b]{0.45\linewidth}
        \includegraphics[width=\linewidth]{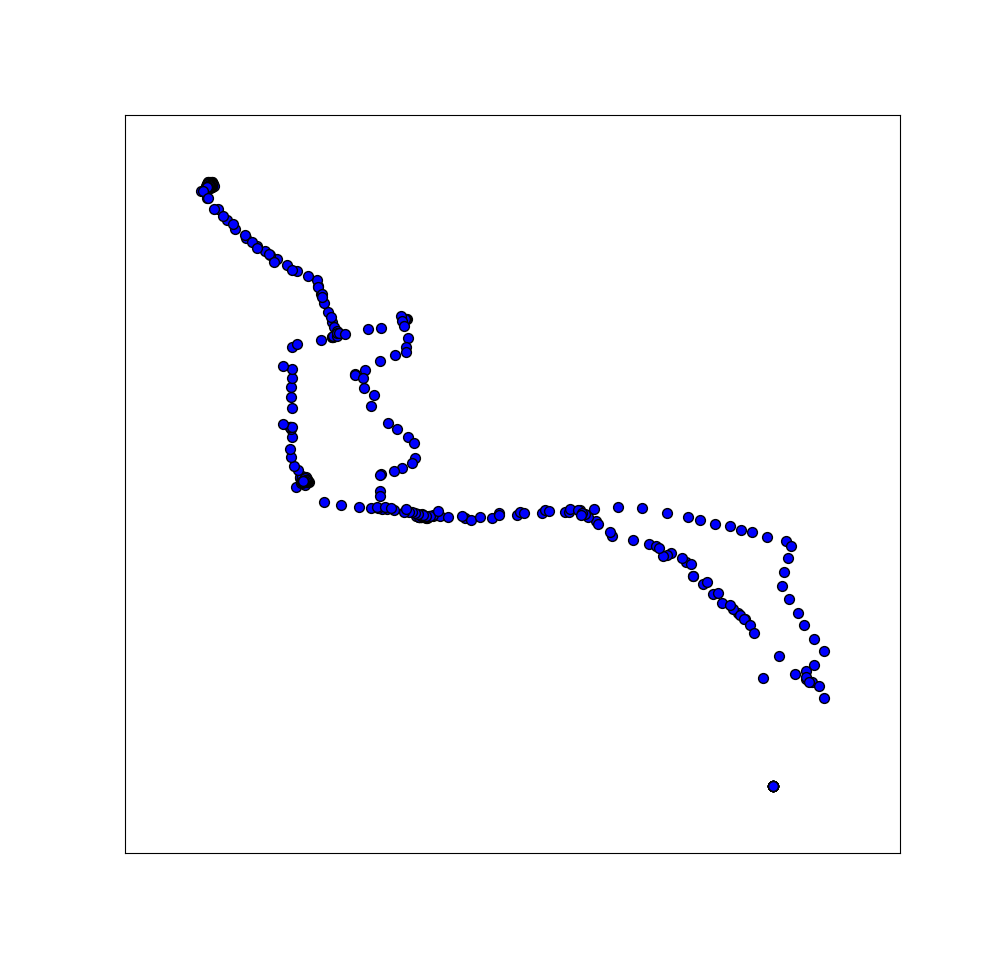}
    \end{subfigure}
    ~ 
    \begin{subfigure}[b]{0.45\linewidth}
        \includegraphics[width=\linewidth]{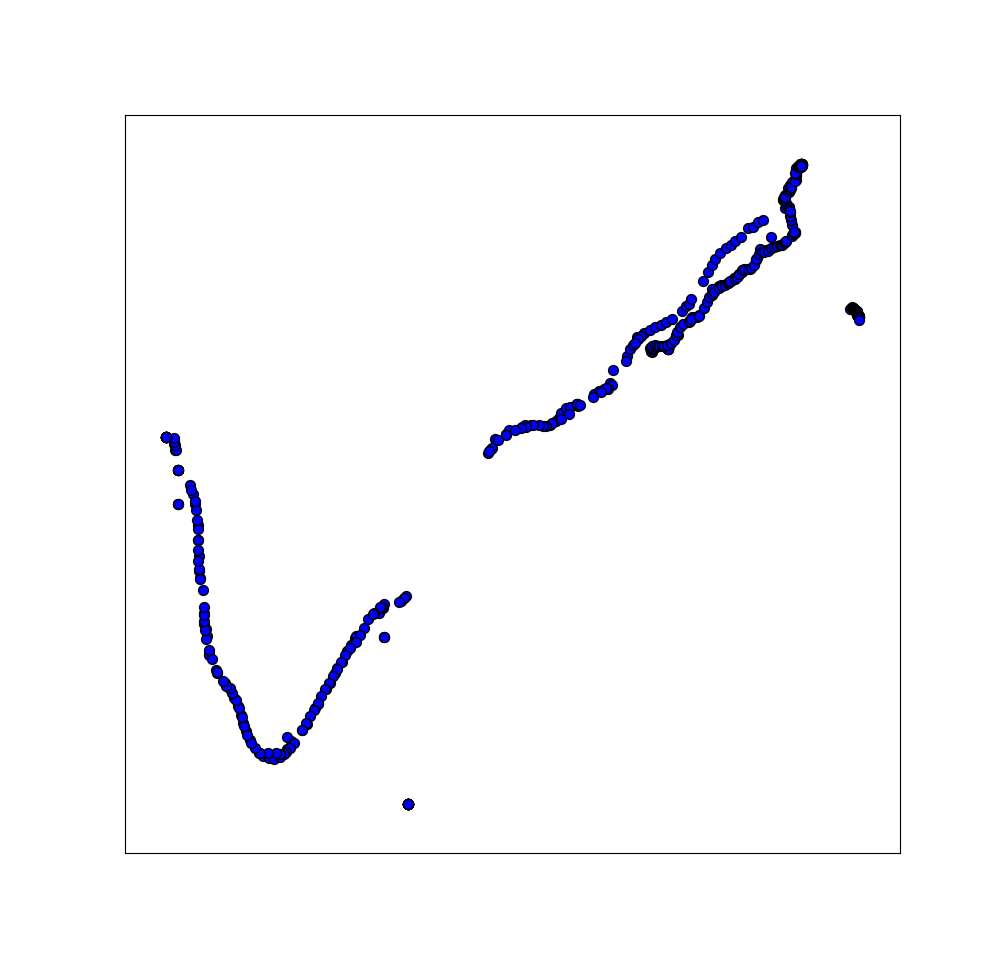}
    \end{subfigure}
    ~     
    
    \caption{Representative trajectories for each discovered mobility patterns.}\label{Fig: Representatives}
\end{figure}

\item \textbf{Number of Patterns and Trajectories}:  Fig.~\ref{Fig: Pattern_num} shows the number of discovered mobility pattern for all the user in our experiments. We can see that the number of mobility patterns varies from $5$ to more than $30$ and most of them are about $10$ to $15$. It also can be found that the lengths of data collecting days are not proportional to the number of discovered mobility patterns, which indicates that the results rely more on the individual behavior rather than the data length.

\begin{figure}[!t]
    \centering
    \includegraphics[width= \linewidth]{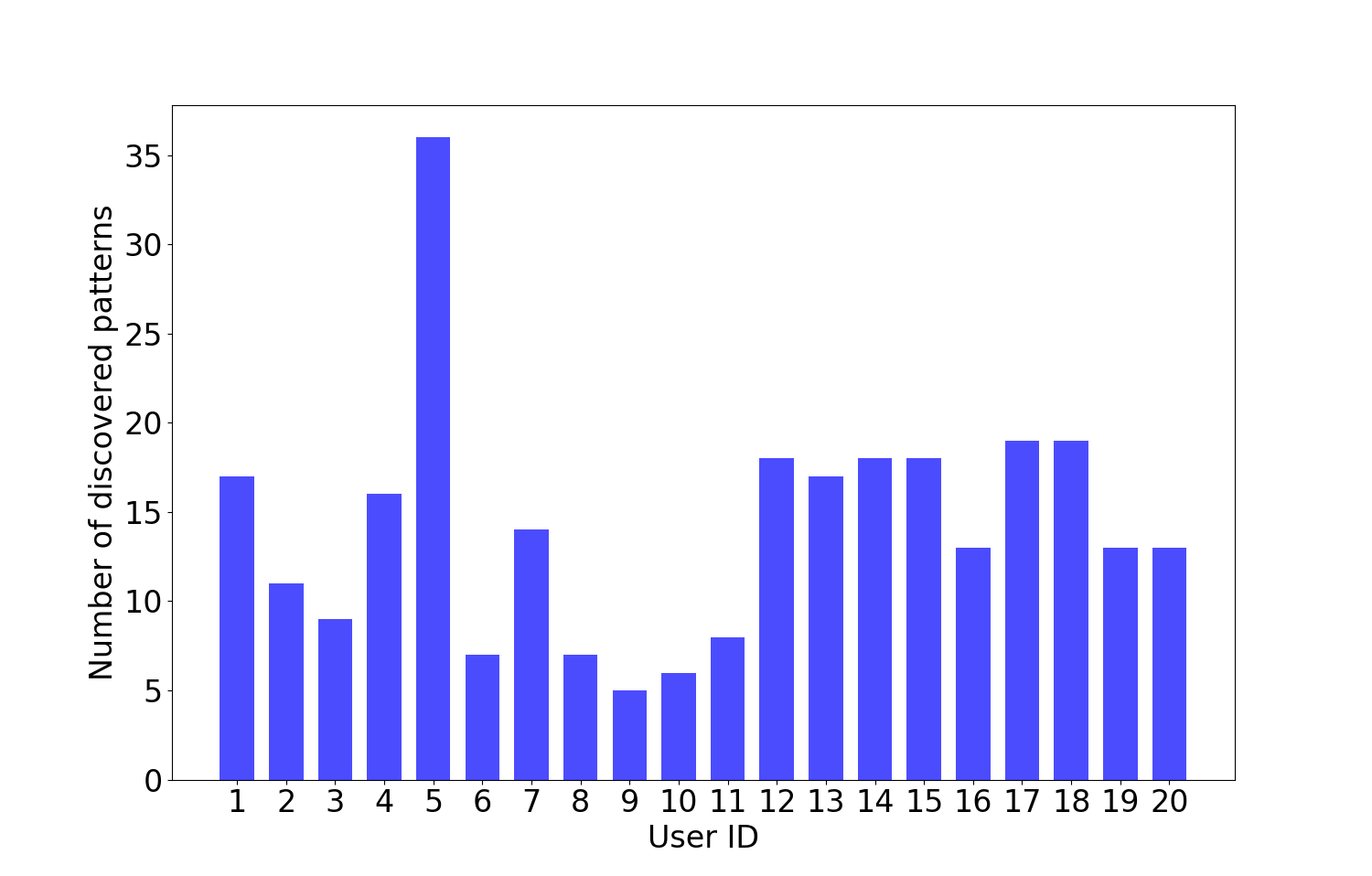}
    \caption{Number of discovered mobility patterns for each user.}\label{Fig: Pattern_num}
\end{figure}

 \item \textbf{Number of members for each patterns}: Fig.~\ref{Fig: Pattern_mem} depicts the number of members for each discovered mobility patterns for all users. We can see that most mobility patterns consist of less than $50$ trajectories. And nearly $40\%$ of the patterns have only one trajectory, whereas few patterns have more than $100$ trajectories.   

 \begin{figure}[!t]
    \centering
    \includegraphics[width= \linewidth]{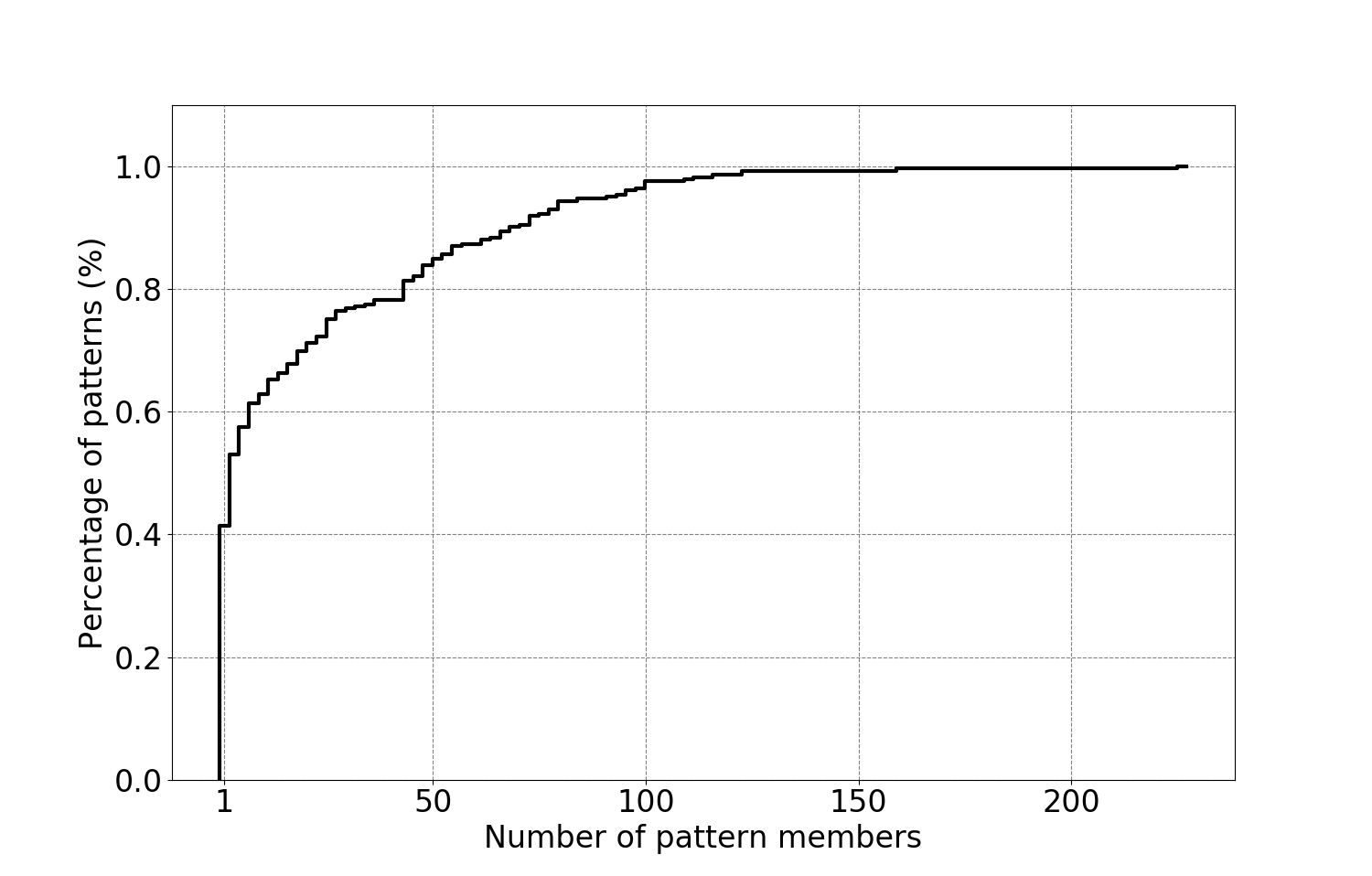}
    \caption{Empirical cumulative distribution of discovered patterns' members.}\label{Fig: Pattern_mem}
\end{figure}

\end{itemize}

One needs to notice that the number of discovered patterns depends on the Kullback-Leibler divergence thresholds we set in the clustering algorithm. When the thresholds are small, it means that the condition to be in the same mobility pattern is more strict and naturally the discovered mobility patterns are more and the member of each patterns are less, and vice versa.

\subsubsection{Comparison to GMM}

To compare the Infinite Gaussian Mixture Models, we use a group of Gaussian Mixture Models with different numbers of components to estimate the daily mobility probability densities in our proposed clustering algorithm. The metrics we adopt to evaluate the results is the mean log-likelihood. The results show in Table~\ref{Table: Log-likelihood Models} that changing the fixed number of component Gaussian Mixture Models can not enhance the clustering performance. On the contrary, the Infinite Gaussian Mixture Models can improve the clustering performance.
 
\captionsetup{font={footnotesize,sc},justification=centering,labelsep=period}%
\begin{table}[htbp]
\caption{Overall Mean Log-likelihood For Different Models}\label{Table: Log-likelihood Models}
\centering%
\begin{tabular}{cc}
\hline
\textit{Model} & \textit{Mean log-likelihood}\\
\hline

GMM-1 & -26078.15\\
GMM-2 & -38514.32\\
GMM-3 & -52431.62\\
GMM-4 & -63794.70\\
GMM-5 & -73508.10\\

IGMM & \textbf{-24871.78}  \\
\hline
\end{tabular}
\end{table}
\captionsetup{font={footnotesize,rm},justification=centering,labelsep=period}%

\subsubsection{Varying Data Length}

\begin{figure}[!t]
    \centering
    \includegraphics[width= \linewidth]{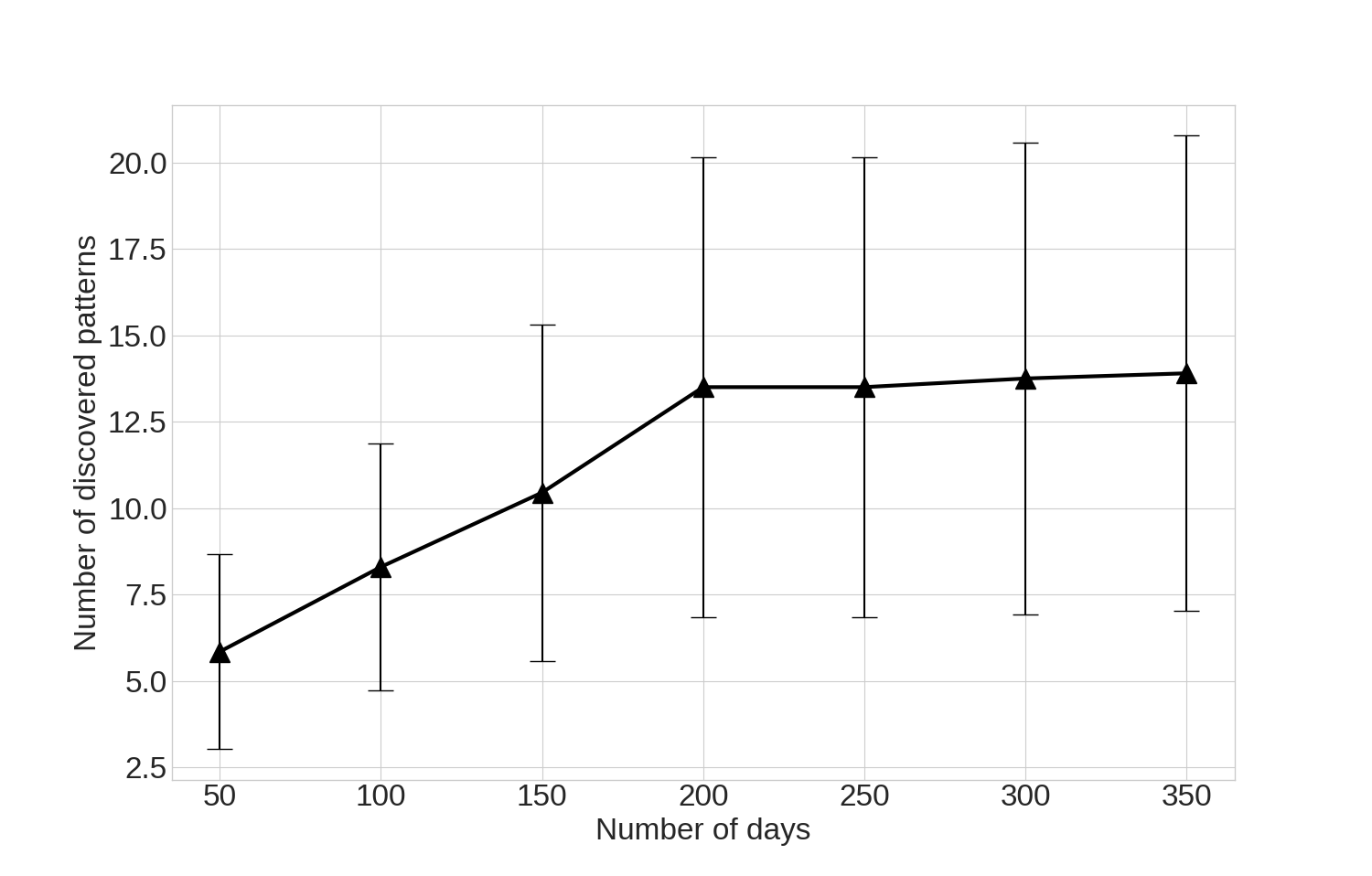}
    \caption{Average number of discovered patterns for different data collecting day length. Error bars represent the standard deviation.}\label{Fig: Pattern_Day}
\end{figure}

To investigate how the data length, namely, the number of days of the data, affects the results, we utilize different data lengths which varies from $50$ days to $350$ days. The results are shown in Fig.~\ref{Fig: Pattern_Day}. It can be seen that, from $50$-day data length to $200$-day data length, the average discovered mobility pattern numbers increase as the data length grows. While, when the data length is larger than $200$ days, the patterns numbers change marginally. Thus, according to the results, we can say that, generally, a $200$-day GPS dataset is large enough to discover most of the mobility patterns of an individual.       

\section{Conclusion and Perspective} \label{Sec: Conclusion and Perspective}

In this work, we presented a probabilistic approach to discover human daily mobility patterns based on GPS data collected by smartphones.

In our approach, the human daily mobility is considered as sets of probability distributions. The proposed approach is divided into three parts. The first step is to estimate the probability densities. We argue that the Infinite Gaussian Mixture Model is more appropriate than the standard Gaussian Mixture Model to this issue, this argument being besides validated by the experimental results. Further, in order to find the similar trajectories, one needs to measure the closeness between the trajectories. For this task, we chose the Kullback-Leibler divergence as distance metrics. According to the computational results from the selected trajectories, we validated, on test sets, that KL divergence is able to measure the similarities among the trajectories. Finally, we devised a novel automatic clustering algorithm combining the advantages of both IGMM and the KL divergence so as to discover human daily mobility patterns without having the knowledge of the cluster number in advance. 

For validation, we select $20$ random individual data from the MDC dataset to conduct the different experiments. The results obtained show that our proposed approach can discern different mobility patterns and select the most representative trajectories for each mobility patterns from the GPS data. In addition, we also compared the IGMM based algorithm with a group of GMM based algorithms with various fix-number components, the results reveal that the IGMM model performs better. Finally, testing varying-length dataset on our methods leads to results which suggest that a $200$-day GPS is generally sufficient enough to discover most of the individual daily mobility patterns.

We are aware of that human mobility is also a time-related behavior. Thus, as future work, we plan to take into account the temporal information, for example, hour of day and day of week. Based on that, we will try to build a spatial-temporal probabilistic model to predict human mobility. In addition, for further study, we may exploit other smartphone usage information (i.g., application usage) in the dataset to obtain more knowledge about human behavior.    

\section*{Acknowledgment}

The research in this paper used the MDC Database made by Idiap Research Institute, Switzerland and owned by Nokia. The authors would like to thank the MDC team for providing the access to the database.


\bibliographystyle{IEEEtran}
\bibliography{bibtemplate_samples}

\end{document}